%% file: main.tex
\documentclass[10pt,twocolumn,letterpaper]{article}

\newif\ifpreprint
\preprinttrue
\newif\ifarxivstyle
\arxivstyletrue
\ifpreprint\else\arxivstylefalse\fi %

\ifpreprint
\usepackage[pagenumbers]{cvpr} %
\else
\usepackage[review]{cvpr}      %
\fi

\input{preamble}
\ifpreprint %
\genmarkfalse\revmarkfalse\fixmarkfalse\shownotesfalse
\fi

\definecolor{cvprblue}{rgb}{0.21,0.49,0.74}
\usepackage[pagebackref,breaklinks,colorlinks,allcolors=cvprblue]{hyperref}

\ifpreprint\else %
\input{xr_setup}
\myexternaldocument{supp}
\fi

\def\confName{CVPR}
\def\confYear{2027}

\title{Scene-Graph Steering for Video Diffusion Models}

\ifpreprint
\newcommand{\projecturl}{https://lilf4p.github.io/graphsteer/}
\author{Leonardo Stoppani$^{1}$ \qquad Donald Shenaj$^{1}$ \qquad Davide Bacciu$^{1}$\\
$^{1}$Department of Computer Science, University of Pisa\\
{\tt\small leonardo.stoppani@phd.unipi.it \quad donald.shenaj@di.unipi.it \quad davide.bacciu@unipi.it}\\[4pt]
{\small\bfseries Project page:}~{\small\href{\projecturl}{\tt lilf4p.github.io/graphsteer}}
}
\else
\author{Anonymous \confName~submission}
\fi

\ifarxivstyle
\usepackage[most]{tcolorbox}
\usepackage{FiraSans} %
\colorlet{headerbg}{oursrow} %

\fi

\begin{document}
\twocolumn[{%
            \renewcommand\twocolumn[1][]{#1}%
            \ifarxivstyle%
  \begin{tcolorbox}[enhanced,colback=headerbg,colframe=headerbg,boxrule=0pt,arc=10pt,
                    left=16pt,right=16pt,top=14pt,bottom=14pt,width=\textwidth]
    {\sffamily\bfseries\LARGE Scene-Graph Steering for Video Diffusion Models\par}
    \vspace{10pt}
    {\sffamily\bfseries\small Leonardo Stoppani$^{1}$, Donald Shenaj$^{1}$, Davide Bacciu$^{1}$\par}
    \vspace{3pt}
    {\small $^{1}$Department of Computer Science, University of Pisa\par}
    {\footnotesize\tt leonardo.stoppani@phd.unipi.it, donald.shenaj@di.unipi.it, davide.bacciu@unipi.it\par}
    \vspace{10pt}
    {\renewenvironment{abstract}{\small\noindent\ignorespaces}{\par}\input{sec/0_abstract}}
    \vspace{8pt}
    {\small\sffamily\bfseries Project page:}~{\small\href{\projecturl}{\tt lilf4p.github.io/graphsteer}}
  \end{tcolorbox}
  \vspace{6pt}\else\maketitle\fi
            \input{sec/teaser}
        }]
\ifarxivstyle\else\input{sec/0_abstract}\fi
\input{sec/1_intro}
\input{sec/2_related}
\input{sec/3_method}
\input{sec/4_experiments}
\input{sec/5_conclusion}

\clearpage
\section*{LLM Usage Statement}
\label{app:llm_usage}
We used the Qwen models for scene graph parsing, baseline layout planning and VLM-based evaluation, as described in the paper. The official T2V-CompBench evaluation of action binding and interaction is done using LLaVA-1.6-34B. Claude was used to assist with code development and debugging. Claude and ChatGPT were used for grammar and language editing and to provide reviewer-style feedback on the final manuscript. This feedback was purely for internal assessment and no LLM generated text from this feedback was used in the manuscript. All LLM assisted content was reviewed by the authors and they take full responsibility for the paper and its results.

    {
        \small
        \bibliographystyle{ieeenat_fullname}
        \bibliography{normalized_main}
    }

\ifpreprint
    \clearpage
    \def\suppinmain{}
    \input{sec/X_suppl}

\fi

\end{document}

%% file: preamble.tex
\usepackage{booktabs}
\usepackage{multirow}
\usepackage{amsmath}
\usepackage{amssymb}
\usepackage{xspace}
\usepackage{xcolor}
\usepackage{colortbl}
\usepackage{graphicx}

\graphicspath{{figures/}}

\definecolor{forestgreen}{RGB}{34,139,34}
\definecolor{lightblue}{RGB}{173,216,230}
\definecolor{brownish}{RGB}
{149, 69, 53}

\definecolor{genblue}{RGB}{0,90,200}
\newif\ifgenmark
\genmarktrue

\definecolor{revmagenta}{RGB}{190,0,140}
\newif\ifrevmark
\revmarktrue

\definecolor{fixteal}{RGB}{0,128,128}
\newif\iffixmark
\fixmarktrue

\newif\ifshownotes
\shownotestrue

\newcommand{\gmn}{Graph Scorer Module\xspace} %
\newcommand{\ours}{GraphSteer\xspace}
\newcommand{\best}[1]{\textbf{#1}}
\definecolor{oursrow}{RGB}{232,240,250}
\newcommand{\oursrow}{\rowcolor{oursrow}}
\definecolor{prefwin}{RGB}{0,130,60}
\definecolor{preflose}{RGB}{200,30,30}
\newcommand{\win}[1]{{\color{prefwin}#1}}
\newcommand{\lose}[1]{{\color{preflose}#1}}

\renewcommand{\paragraph}[1]{\vspace{.4em}\noindent\textbf{#1}}

%% file: sec/0_abstract.tex
\begin{abstract}
Text-to-video diffusion models can generate remarkable videos,
but still struggle when prompts involve multiple relations between
objects, often missing or misplacing some of them.
Existing approaches tackle this problem using box-based guidance
to enforce the desired relations. However, this requires defining
the scene layout in advance, manually or using an LLM, for every
prompt. This makes the process impractical when scaling to more
complex scenes, as it relies on a fixed layout that does not
account for what the generator actually produces.
We address these limitations by introducing \mbox{\textbf{\ours}},
a plug-and-play method that uses scene graphs to guide
generation towards satisfying multiple relations,
without requiring any predefined layout.
\ours represents the prompt as a scene graph, 
whose nodes are objects and whose edges describe
their spatial relations.
We design a \gmn that scores every edge directly from
cross-attention, so all relations are optimized through
a single joint loss.
We also observe that existing text-to-video benchmarks are not
designed to evaluate multiple relations jointly, making it
difficult to identify individual relation failures or assess
whether all relations hold in the same video.
To address this, we introduce multi-relation prompts and a global
preference score based on pairwise VLM judgments.
On CogVideoX-5b, \ours improves the T2V-CompBench spatial score
from $0.545$ to $0.631$ and multi-relation per-edge accuracy
from $0.225$ to $0.392$.
Finally, we show that \ours transfers zero-shot to different
video generation 
architectures, including Wan2.2-A14B and
HunyuanVideo, without retraining.
\end{abstract}

%% file: sec/teaser.tex
\begin{center}
\includegraphics[width=0.99\linewidth]{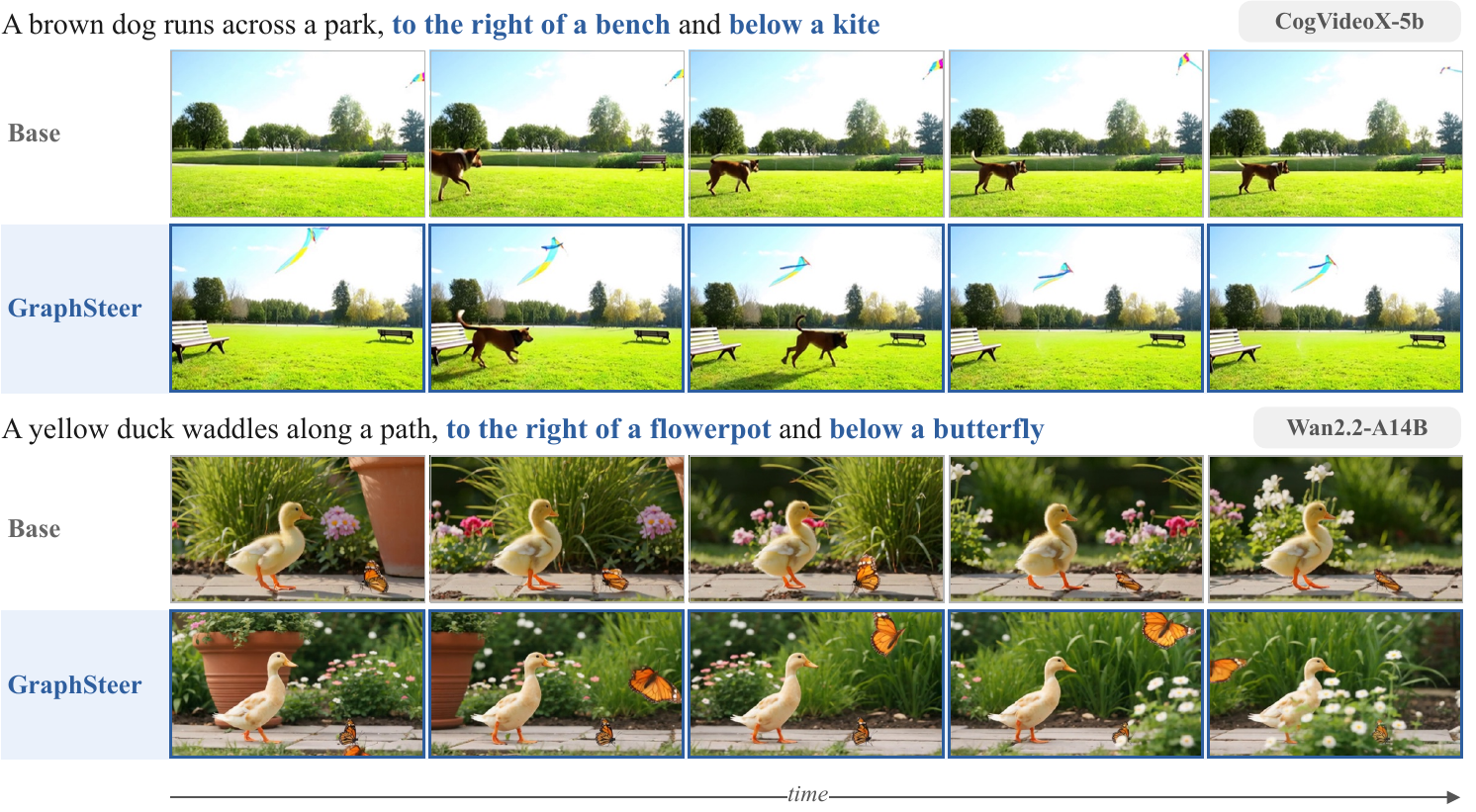}
    \captionof{figure}{
We introduce \textbf{\ours}, which steers frozen video backbones towards all the relations
in the prompt, without layouts or retraining.
Top (CogVideoX-5B): the base places the dog left of the bench, while
\ours makes it run to the right of the bench, underneath the kite.
Bottom (Wan2.2-A14B): the base loses the flowerpot and keeps the butterfly
on the ground, while with \ours the flowerpot stays on the left and the butterfly
takes off above the duck, using a \gmn never trained on Wan attention.
    }
  \label{fig:teaser}
\end{center}

%% file: sec/1_intro.tex
\section{Introduction}
\label{sec:intro}

Text-to-video (T2V) diffusion models can now create sharp and temporally coherent clips
\cite{yang2024cogvideox,wan2025,kong2024hunyuanvideo}, but they still frequently fail in spatial arrangements. For example, an object might land on the wrong
side of another object, or it might not appear at all (\cref{fig:teaser}).
Established benchmarks measure these failures \cite{sun2024t2vcompbench}, which occur even more often when
a prompt includes several objects with multiple relations that must hold simultaneously.

Recent approaches to T2V generation address this challenge by dividing the task into two stages. Initially, a layout planner determines the spatial and temporal coordinates for subject placement. This information is then provided to the generator, which utilizes it by (i) optimizing the latent representation against an attention energy \cite{lian2024llmgroundedvideo}; (ii) directly editing attention maps during sampling \cite{tian2024videotetris,zhang2025compattn}; or (iii) fitting a LoRA to each video \cite{qu2026ttom}.
A common limitation of these methods is that spatio-temporal reasoning is delegated to an external planner that sees only the text prompt and never observes what the generator produces. The resulting layouts are fixed before sampling, may conflict with the generator’s prior, and convert relative relations into rigid absolute boxes. Each object is then constrained independently, so nothing ensures that all relations hold jointly.

A scene graph represents an image or video as a set of objects, called nodes, connected by relationships. These relationships, known as edges, define the connections between nodes \cite{johnson2015retrieval,krishna2017visualgenome}. They provide a more comprehensive description of a scene than a simple list of bounding boxes, as they can represent any number of objects, and each object can be part of several relationships. This approach does not specify an object's absolute position or size; instead, it uses relative terms such as ``left of" or ``in front of." Scene graphs have been applied to image retrieval and generation \cite{johnson2015retrieval,johnson2018sg2im,shen2024sgadapter} and, in video, mostly to recognition \cite{shang2017vidvrd,ji2020actiongenome}. The choice of using scene graphs is central to \ours. Relationships between objects can be evaluated directly on the generator's output without explicit spatial planning. Moreover, because edges share nodes, we can evaluate each relationship jointly with others, enabling us to enforce all relationships with a single loss function over the entire graph.

We introduce \ours,
which enforces a frozen T2V generator to capture all the relationships jointly by acting directly on its attention maps, without layouts, planners, or retraining.
First, a prompt is parsed into a scene graph describing entities and their relations in the video. Then every relation is scored from the attention maps of each entity, and a single loss is applied across the entire graph to the early denoising steps of a frozen T2V diffusion model. We resample attention onto a fixed grid, which enables zero-shot transfer to 
different DiT-based \cite{peebles2023dit} video generators. 

We evaluate our approach against existing layout-based methods on standard compositional benchmarks and study whether multiple relations can be satisfied jointly in the same video. We also investigate how steering affects the overall generation quality.
Although many benchmarks have been proposed for controllable T2V generation \cite{huang2024vbench,liu2024evalcrafter,sun2024t2vcompbench}, current evaluation protocols suffer from two issues: they cannot verify whether multiple relations hold simultaneously within a single clip, and averaging independent per-category scores can favour methods tuned to individual categories rather than methods that improve generation as a whole.
To address the former, we design a set of multi-relation prompts. To address the latter, we propose a global preference score: inspired by the LLM-as-a-judge paradigm~\cite{zheng2023judging}, a VLM compares two clips in both presentation orders to mitigate position bias~\cite{wang2024fairevaluators}.

The main contributions of our work are:
\begin{itemize}
    \setlength\itemsep{0.1em}
    \item \textbf{Scene-graph relation steering:}
    We introduce a \gmn that scores multiple relations directly from cross-attention maps and guides a frozen video generator using a single joint loss (\cref{sec:method}). Unlike prior work, we neither condition the generator on the scene graph nor turn the graph into a fixed layout: the graph only defines a relational loss that steers a frozen model at inference time, and the \gmn is trained on video scene-graph annotations without retraining the generator.

    \item \textbf{Layout-free relational control across video generators:}
    On CogVideoX-5b, \ours improves the T2V-CompBench spatial score from $0.545$ to $0.631$. The same module transfers to HunyuanVideo without retraining, improving the spatial score from $0.610$ to $0.715$ (\cref{sec:exp_compbench,sec:exp_wan}).

    \item \textbf{Multi-relation evaluation and global preference:}
    We introduce a multi-relation benchmark (\cref{sec:exp_multirel}) to evaluate individual relation correctness and a VLM-based global preference score (\cref{sec:exp_quality}) to compare how well multiple relations are satisfied in the same video.
\end{itemize}

%% file: sec/2_related.tex
\section{Related work}
\label{sec:related}

\begin{figure*}[t]
  \centering
  \includegraphics[width=0.97\linewidth]{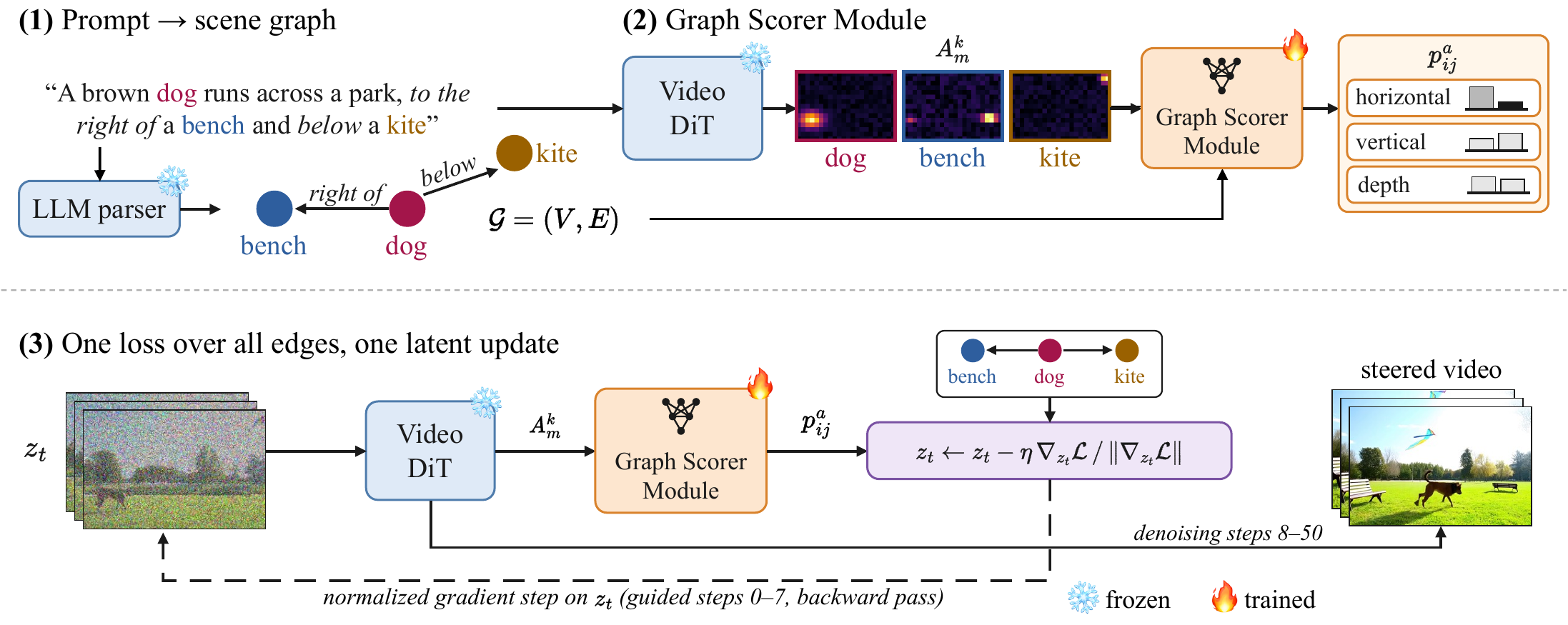}
  \caption{Overview of the \ours pipeline. (1) An LLM parses the prompt into a scene graph with per-axis edge targets. (2) The \gmn scores every edge on each named axis from the entity attention of the frozen video DiT, resampled to a resolution-free grid, giving the probabilities $p^a_{ij}$. (3) One loss over all edges (\cref{eq:loss}) compares these scores with the scene-graph targets; its normalized gradient updates $z_t$ during the first 8 denoising steps, then sampling continues unchanged. The generator stays frozen.}
  \label{fig:method}
\end{figure*}

\paragraph{Inference-time steering.}
Attention-based losses can steer a frozen model toward object presence, attribute binding and timing \cite{chefer2023attendexcite,rassin2023syngen,feng2023structurediffusion,meral2024conform,li2025starvid,schiber2026tempocontrol}, and universal guidance follows the gradient of any differentiable objective \cite{bansal2023universal}.
Learn-to-Steer \cite{learn2steer} learns the steering objective directly from cross-attention maps, training a classifier to recognize spatial relations between pairs of objects in text-to-image models. It also handles multiple relations, but optimizes them separately by alternating the target relation during denoising. Differently, we target video generation and introduce a graph-based objective that scores all relations of a scene graph and optimizes them jointly through a single loss.

\paragraph{Layout-conditioned generation.}
Most compositional methods plan a box layout, often with an LLM, and make the generator follow it through grounding layers trained on boxes \cite{li2023gligen}, box-constrained attention \cite{xie2023boxdiff}, or attention energies and edits at sampling time \cite{lian2024llmgroundedvideo,tian2024videotetris,zhang2025compattn}. TTOM \cite{qu2026ttom} also optimizes a LoRA \cite{hu2022lora} at inference time, while VideoComposer and Direct-a-Video use additional motion or spatial controls \cite{wang2023videocomposer,yang2024directavideo}, as do trajectory-based methods \cite{wu2024draganything,li2025magicmotion,fu2024trajmaster}.
These methods rely on an external planner that defines the layout before generation, without knowing what the generator will actually produce. The resulting layout is fixed and may conflict with the generator's prior. Moreover, 2D bounding boxes only specify object positions, without telling which object should appear in front when two objects overlap. 
Differently, we guide generation directly from the relations in the prompt, without requiring any predefined layout.

\paragraph{Scene graphs for generation.}
Existing scene-graph-to-image methods train a generator or an adapter to generate images conditioned on a graph \cite{johnson2018sg2im,yang2022sg2im,liu2024r3cd,shen2024sgadapter,wu2024relationrectification,wang2024disco}. Other approaches follow a two-step process, first converting the text into a layout using an LLM and then using it to guide generation, as in LayoutLLM-T2I \cite{qu2023layoutllmt2i}. In video, GraphVid \cite{shah2026graphvid} trains an image-to-video model conditioned on interaction graphs, while scene graphs have otherwise mainly been used for recognition \cite{shang2017vidvrd,shang2019vidor,ji2020actiongenome,liu2025openvidvrd}. In this work, we take a further step towards layout-free relational control in video generation. Instead of using a graph to condition the generator or define a fixed layout, we use it to build a joint relational loss that guides a frozen model at inference time. We achieve this through a \gmn trained on video annotations, without retraining the video generator.

%% file: sec/3_method.tex
\section{\ours}
\label{sec:method}

\ours introduces a graph-based steering objective to guide a frozen T2V diffusion model during denoising, without requiring any predefined layout or retraining (\cref{fig:method}). Given a text prompt, we first construct a scene graph describing the entities and their relations. We then use a learned \gmn to score all the relations directly from the entities' cross-attention maps (\cref{sec:classifier}). These scores are combined into a single loss that guides the latent during the early denoising steps (\cref{sec:guidance}), while keeping the generator's weights unchanged.

We represent each prompt as a directed graph $G=(V,E)$, where nodes correspond to the entities in the prompt and edges describe their spatial relations. Each relation is associated with a geometric axis and a sign, indicating the relative position of two entities. Although several scene graph parsers exist \cite{schuster2015sgparser,wu2019sgparser,li2023factual,honnibal2020spacy}, we use an open-source LLM \cite{qwen2025} to handle different ways of expressing the same spatial relation. We first extract the entities and their corresponding tokens to construct the node set $V$, and then represent each relation as a subject, object, axis, and sign. An axis is one of the three directions along which we compare two entities: horizontal (left/right), vertical (above/below), and depth (in front/behind). On each axis, the sign tells which of the two opposite relations holds, for example whether the subject is to the left or to the right of the object. More formally, an edge $e=(i,j)\in E$ from subject $i\in V$ to object $j\in V$ carries a target sign $r^\star_{e,a}\in\{-1,+1\}$ for each axis $a$ in the set $\mathcal{A}(e)$ of axes the prompt specifies for it, out of horizontal, vertical, and depth. A relation such as ``above and in front of'' constrains two axes, and a relation such as ``left of'' only one.
Using gold entities, the LLM parser achieves exact-graph
agreement on 99.2\% of our multi-relation prompts and
97.5\% of T2V-CompBench spatial prompts (Suppl. \cref{app:parser}).

\subsection{\gmn}
\label{sec:classifier}
The \gmn is block (2) of \cref{fig:method}. It encodes the attention maps of each entity into a node feature, and a graph neural network then scores every edge of the scene graph from these features.

\paragraph{Attention features.}
Attention scores between video and text tokens localize prompt concepts \cite{hertz2022prompt}. Steering methods use this information by optimizing for spatial fidelity, defining a loss over the entities' attention maps \cite{chefer2023attendexcite,learn2steer}. 

For each entity $k\in V$, we average over its tokens in each head $m\in\{1,\dots,M\}$ of one transformer block, yielding maps $A^k_m\in\mathbb{R}^{F\times H\times W}$ over the grid of video tokens ($F$ latent frames of $H\times W$ tokens). Unlike image-based approaches, we preserve the temporal dimension of the attention maps, allowing the \gmn to evaluate object relations across video frames rather than on a single image.
The choice of extraction block depends on the backbone model; we detail this in the supplementary material. Backbones also differ in grid size, head count, and attention scale. We handle these differences by resampling each map to $8{\times}16{\times}16$, normalizing it by its own mass, and processing the resulting maps as an unordered set. A small transformer encodes these maps as tokens, and a learned query pools them into a node feature $u_k$. This allows the same encoder weights to handle different numbers of attention heads across generators.

\paragraph{Edge scoring.}
Each $u_k$ passes through two message-passing layers of a Graph Neural Network (GNN) with residual connections \cite{brody2022gatv2}, producing a contextual embedding $h_k$. For each edge $e=(i,j)$, a shared encoder processes $h_i$, $h_j$, $h_i{-}h_j$, the mean of $h_k$ over all nodes, and statistics about the two attention maps. Subsequently, one softmax per spatial axis (horizontal, vertical, depth) outputs $p^a_{ij}$, a probability distribution over the two signs of axis $a$, as shown in \cref{fig:method}.
This allows multiple axes to fire at the same time,
consistent with the nature of video data observed.
On ternary relations, 
graph message
passing substantially improves steering over independent
pairwise scoring (Suppl. \cref{app:between}).

\paragraph{Attention maps extraction.}
To train the \gmn, we extract attention maps from real and synthetic video clips. Following DIFT \cite{tang2023dift}, we first forward-noise the clean video latent $z_0$ as
$z_t=\sqrt{\bar\alpha_t}z_0+\sqrt{1-\bar\alpha_t}\,\epsilon$, where $\epsilon\sim\mathcal{N}(0,I)$ and $\bar\alpha_t$ is the cumulative noise schedule,
at step $t=5$, and run a single forward pass to obtain the attention maps. This step lies inside the window we later guide (the first $8$ as shown in \cref{fig:method}), so the \gmn is trained on maps from the same range of noise levels it reads during steering. We process each clip using both its original prompt and a corrupted version, keeping the same ground-truth relations in both cases. This prevents the \gmn from relying only on the prompt rather than the visual content \cite{learn2steer}.

\subsection{Joint guidance over the graph}
\label{sec:guidance}

\begin{figure}[t]
  \centering
  \includegraphics[width=\linewidth]{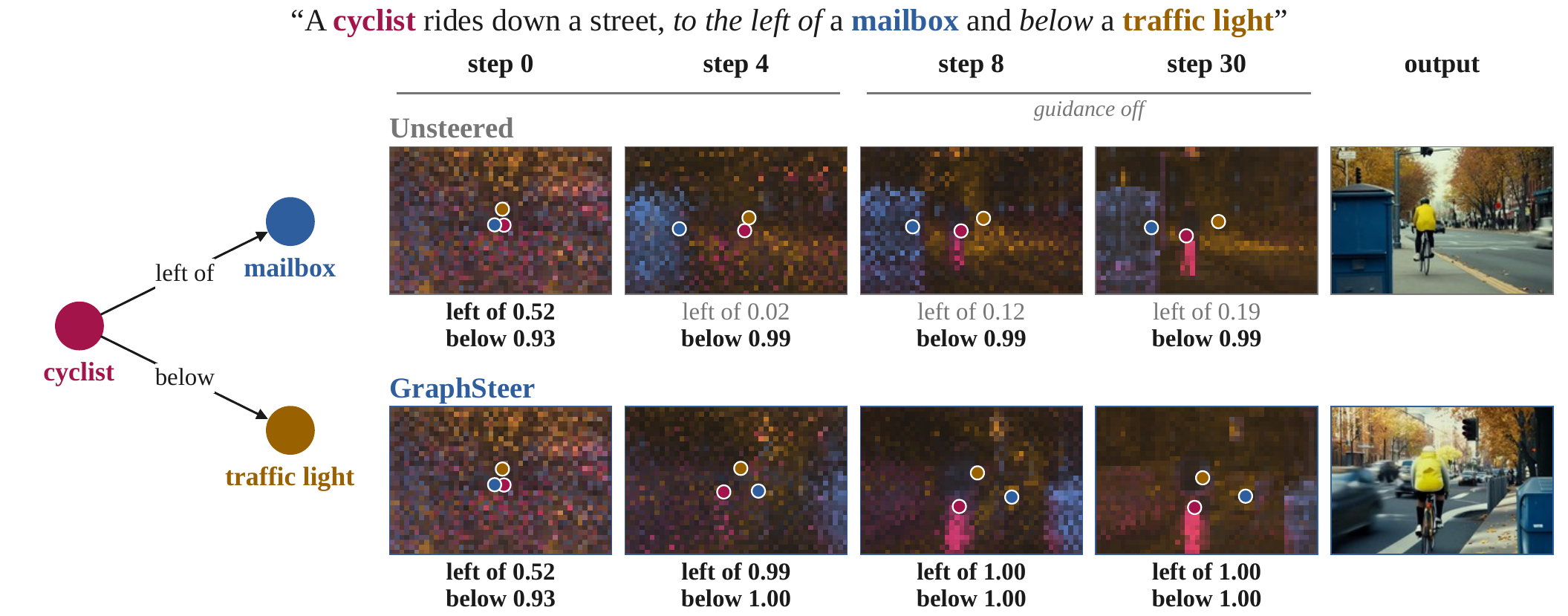}
  \caption{
  Entity attention maps with and without \ours on
CogVideoX-5b, for the prompt and scene graph (left).
Each tile shows head-averaged cross-attention for three
entities at the \gmn's input block, on the middle latent
frame. Dots mark attention centroids; numbers show the
\gmn's probabilities of the requested relations.
Both rows share the prompt and seed.
  }
  \label{fig:attn_steer}
\end{figure}

Let $z_t$ be the video latent at denoising step $t$. At each guided step, we run one forward pass
of the generator's text-conditioned branch with gradients enabled, extract the entity maps $A^k_m$,
and feed them to the \gmn, which returns $p^a_{ij}(z_t)$ for every edge and axis.
We write $p^a_{ij}(z_t)[r]$ for the probability it assigns to sign $r\in\{-1,+1\}$, that is, to one of the two opposite relations on axis $a$ (\eg, left of or right of). The loss of an edge $e=(i,j)$
is the cross-entropy against the target signs $r^\star_{e,a}$ parsed from the prompt, averaged over the axes of that edge, and
the relation loss sums it over all edges:
\begin{align}
  \ell_e(z_t) = -\frac{1}{|\mathcal{A}(e)|}\sum_{a\in\mathcal{A}(e)}
     \log p^{a}_{ij}(z_t)\big[r^\star_{e,a}\big], \label{eq:edge}\\
  \mathcal{L}_{\mathrm{rel}}(z_t)= \sum_{e\in E}\ell_e(z_t). \label{eq:loss}
\end{align}
Averaging over axes keeps a pair constrained on two axes from pulling twice as hard as a pair
constrained on one, and summing over edges lets every relation of the prompt shape the same gradient.

Each map is divided by its own mass before it reaches the network (\cref{sec:classifier}), so the
network cannot tell how much attention an entity receives, and the gradient can satisfy a relation by
erasing an object. We therefore track the raw attention mass of each entity before normalization,
$m_k(z_t)=\sum_{m}\sum_{f,x,y}A^k_m[f,x,y]$, and penalize large drops, with respect to its value
$m_k^0$ at the first guided step:
\begin{equation}
  \mathcal{L}_{\mathrm{pres}}(z_t)=\sum_{k\in V}\max\!\Big(0,\ \frac12-\frac{m_k(z_t)}{m_k^{0}}\Big).
  \label{eq:pres}
\end{equation}
Without this term ($\lambda{=}0$), steering still improves over the base
model, but T2V-CompBench spatial drops from 0.631 to 0.590 and per-edge
accuracy from 0.392 to 0.366 (Suppl. \cref{app:presence}).
With $\mathcal{L}=\mathcal{L}_{\mathrm{rel}}+\lambda\,\mathcal{L}_{\mathrm{pres}}$,
where $\lambda$ weights the presence term, the latent takes a gradient step of fixed length $\eta$:
\begin{equation}
  z_t \leftarrow z_t - \eta\,\frac{\nabla_{z_t}\mathcal{L}}{\lVert\nabla_{z_t}\mathcal{L}\rVert}.
  \label{eq:update}
\end{equation}
Since the gradient is normalized, $\eta$ sets the length of every update regardless of the scale of the loss. A single update moves the latent towards every relation of the prompt at once, because $\mathcal{L}_{\mathrm{rel}}$ sums the losses of all edges (\cref{eq:loss}). We guide the first 8 of 50 steps as a trade-off between accuracy and
cost (Suppl. \cref{app:steps}). \cref{fig:attn_steer} shows our method in action on one prompt with CogVideoX-5b as base. Both runs start from the same latent, so the \gmn reads the same maps at step $0$ and gives ``left of'' a probability of $0.52$. Without guidance, the cyclist's map ends up to the right of the mailbox's, and the probability of the requested relation is $0.02$ at step $4$, $0.12$ at step $8$ and $0.19$ at step $30$. With guidance it is $0.99$ at step $4$ and $1.00$ at steps $8$ and $30$: the mailbox map moves to the right of the cyclist's, and the traffic light's map stays above both. The ``below'' edge already holds in both runs, and the updates do not disturb it. Guidance acts on steps $0$ to $7$ only, and the relation still holds at step $30$, so the arrangement reached in that window persists. These probabilities come from the same \gmn that defines the loss.

%% file: sec/4_experiments.tex
\section{Experiments}
\label{sec:experiments}

In this section, we evaluate \ours against existing
approaches to relational control in T2V generation.

\paragraph{Settings.}
Following prior work~\cite{lian2024llmgroundedvideo,tian2024videotetris,
qu2026ttom,zhang2025compattn,sun2024t2vcompbench},
we report one run per prompt with a fixed seed.
Within each backbone, our runs of base, LVD, Comp-Attn,
and \ours share the same prompt, seed, and initial noise
(Suppl.~\cref{app:statistical_reporting}).
All \ours video results use the same \gmn checkpoint
and guidance settings across backbones, changing only
the attention extraction block.
On Wan2.2-A14B, all guided steps fall on the high-noise expert.
Supplementary material lists frames and resolutions.
We train the \gmn using two video datasets annotated with scene graphs, VidOR and ImageNet-VidVRD \cite{shang2019vidor,shang2017vidvrd},
plus approximately $600$ synthetic clips annotated using GroundingDINO
bounding boxes \cite{liu2024groundingdino} rather than their generation prompts.

\paragraph{Competitors.}
We compare against VideoTetris \cite{tian2024videotetris}, TTOM \cite{qu2026ttom}, LVD, and Comp-Attn.
For VideoTetris and TTOM, we follow their released pipelines, replacing closed-source planners with Qwen models.
Qwen2.5-VL-7B \cite{qwen2.5-VL} satisfies $98.3\%$ of VideoTetris's layout relations.
For TTOM, Qwen2.5-72B achieves layout scores close to GPT-4o ($0.906$ vs.\ $0.915$)
and reproduces its T2V-CompBench spatial score ($0.634$ vs.\ $0.632$; Suppl. \cref{app:competitors}).
We also use Qwen2.5-72B for layout planning in LVD and Comp-Attn,
running both on the same backbones and seeds.

\paragraph{Metrics.}
We use T2V-CompBench to evaluate standard compositional properties
(\cref{sec:exp_compbench}).
Since T2V-CompBench considers only one relation per prompt, while real prompts can involve several relations between the same objects,
we introduce a multi-relation benchmark where a shared object must
satisfy several relations in the same video.
We measure per-edge accuracy to assess how many relations are correctly
generated (\cref{sec:exp_multirel}).

\subsection{T2V-CompBench}
\label{sec:exp_compbench}

We first evaluate whether \ours can improve spatial relations without relying on a predefined layout, comparing it against methods that explicitly plan one. On CogVideoX-5b \cite{yang2024cogvideox}, \ours gives the largest gain in T2V-CompBench \cite{sun2024t2vcompbench} spatial score among the methods of \cref{tab:competitors} ($+0.086$), ahead of LVD and Comp-Attn on the same backbone and of TTOM on Wan2.1, even though it is the only one that receives no layout. Numeracy also improves, and the other five dimensions stay within $0.02$ of the base model (Suppl. \cref{app:compbench7}). On Wan2.2-A14B, where \ours runs zero-shot, the spatial gain is smaller and LVD scores best, but \ours is the only method in that block that does not lower numeracy. We return to transfer in \cref{sec:exp_wan}.

\begin{table}[t]\centering\small
  \setlength{\tabcolsep}{3pt}
  \resizebox{\linewidth}{!}{%
  \begin{tabular}{llcc}
    \toprule
    method & layout input & Spatial ($\Delta$) & Numeracy ($\Delta$) \\
    \midrule
    VideoTetris \cite{tian2024videotetris} & LLM boxes & 0.483 & 0.347 \\
    TTOM$^\dagger$ \cite{qu2026ttom} & LLM boxes + LoRA & 0.632 ($+0.063$) & 0.478 ($-0.024$) \\
    \midrule
    CogVideoX-5b                   & ---           & 0.545 & 0.278 \\
    \quad + ViCo \cite{yang2024vico} & ---         & 0.545 & \best{0.395} \\
    \quad + LVD$^\ast$ \cite{lian2024llmgroundedvideo} & LLM boxes & 0.504 ($-0.040$) & 0.271 ($-0.008$) \\
    \quad + Comp-Attn$^\ddagger$ \cite{zhang2025compattn} & LLM boxes & 0.601 ($+0.057$) & 0.348 ($+0.070$) \\
    \oursrow\quad + \ours                  & \textbf{none} & \best{0.631} ($+0.086$) & 0.346 ($+0.068$) \\
    \midrule
    Wan2.2-A14B                    & ---           & 0.652 & 0.594 \\
    \quad + ViCo                   & ---           & 0.673 & 0.503 \\
    \quad + LVD$^\ast$             & LLM boxes     & \best{0.703} ($+0.051$) & 0.558 ($-0.035$) \\
    \quad + Comp-Attn$^\ddagger$   & LLM boxes     & 0.667 ($+0.015$) & 0.548 ($-0.045$) \\
    \oursrow\quad + \ours (zero-shot)      & \textbf{none} & 0.674 ($+0.022$) & \best{0.597} ($+0.004$) \\
    \bottomrule
  \end{tabular}}
  \caption{\textbf{T2V-CompBench} spatial and numeracy.
$\Delta$ is the gain over our run of the same backbone and seed.
Bold marks the best within each backbone block; shaded rows are ours.
Rows without $\Delta$ are published scores: VideoTetris on
VideoCrafter2 \cite{sun2024t2vcompbench} and ViCo as reported
in \cite{zhang2025compattn}.
$^\dagger$TTOM reproduced on Wan2.1-14B with GPT-4o layouts.
$^\ast$Adapted. $^\ddagger$Reimplemented.
Baseline details in Suppl. \cref{app:competitors}.
}
  \label{tab:competitors}
\end{table}

\begin{table*}[t]\centering\footnotesize
  \setlength{\tabcolsep}{5pt}
  \resizebox{\textwidth}{!}{%
  \begin{tabular}{llcccccc|c}
    \toprule
    & & \multicolumn{3}{c}{per-edge accuracy by relation} & & & & \\
    \cmidrule(lr){3-5}
    method & layout input & left/right$\uparrow$ & above/below$\uparrow$ & front/behind$\uparrow$ & per-edge$\uparrow$ & all-rel.$\uparrow$ & count$\uparrow$ & GPU-min / clip$\downarrow$ \\
    \midrule
    VideoTetris \cite{tian2024videotetris} & LLM boxes & 0.291 & 0.386 & 0.284 & 0.307 & 0.101 & 0.566 & 1.0$^\P$ \\
    TTOM \cite{qu2026ttom}     & LLM boxes + LoRA & 0.344 & 0.727 & 0.206 & 0.375 & 0.100 & 0.874 & 47$^\P$ \\
    \midrule
    CogVideoX-5b               & ---           & 0.125 & 0.500 & 0.235 & 0.225 & 0.058 & 0.534 & 4.2 \\
    \quad + LVD$^\ast$         & LLM boxes    & 0.156 & 0.455 & 0.088 & 0.192 & 0.058 & 0.432 & 11.7$^\P$ \\
    \quad + Comp-Attn$^\ddagger$   & LLM boxes    & \best{0.383} & 0.523 & \best{0.279} & 0.379 & 0.133 & 0.634 & 9.8$^{\P\S}$ \\
    \oursrow\quad + \ours              & \textbf{none} & 0.367 & \best{0.636} & \best{0.279} & \best{0.392} & \best{0.150} & \best{0.740} & 5.3 \\
    \midrule
    Wan2.2-A14B                & ---          & 0.195 & 0.455 & 0.147 & 0.229 & 0.017 & 0.758 & 17.2 \\
    \quad + LVD$^\ast$         & LLM boxes    & \best{0.492} & \best{0.636} & \best{0.162} & \best{0.425} & \best{0.158} & \best{0.784} & 33.5$^\P$ \\
    \quad + Comp-Attn$^\ddagger$   & LLM boxes    & 0.273 & 0.409 & 0.088 & 0.246 & 0.033 & 0.700 & 21.8$^\P$ \\
    \oursrow\quad + \ours (zero-shot)  & \textbf{none} & 0.258 & 0.477 & \best{0.162} & 0.271 & 0.033 & 0.707 & 19.0 \\
    \bottomrule
  \end{tabular}}
  \caption{
  \textbf{Multi-relation benchmark.} 
        Bold marks the best result within each backbone block;
        shaded rows are ours. Count follows the T2V-CompBench
        numeracy rule. GPU cost (lower is better) includes model
        loading and planning, measured in GPU-minutes on A100-64GB GPUs.
        TTOM runs on Wan2.1-14B and VideoTetris on VideoCrafter2.
        $^\ast$Adapted. $^\ddagger$Reimplemented. Both are our runs on the same base and seed.
        $^\P$Includes layout planning.
        $^\S$Additional attention computation for Comp-Attn.
        Details in Suppl. \cref{app:baseline_costs}.}

  \label{tab:multirel}
\end{table*}

\subsection{Multi-relation benchmark}
\label{sec:exp_multirel}

We now investigate whether several relations involving the
same objects can be satisfied in a single video.

\paragraph{Benchmark.}
We construct a benchmark of $120$ natural scenes, each containing three objects and two spatial relations sharing one object, with a ground-truth
graph used only for evaluation.
Following T2V-CompBench scoring protocol \cite{sun2024t2vcompbench}, we use GroundingDINO \cite{liu2024groundingdino} bounding boxes for left/right and above/below relations, and depth estimates within SAM masks for front/behind (Suppl. \cref{app:multirel}). 
We report per-edge accuracy, the fraction of correctly generated
relations, and all-relations rate, the fraction of videos where
both relations hold simultaneously.

\paragraph{Results.}
On CogVideoX-5b, \ours achieves the best per-edge accuracy,
all-relations rate, and object count (\cref{tab:multirel}).
Although VideoTetris plans boxes satisfying $98.3\%$ of
the requested relations, only about one edge in three
holds in its generated clips.
TTOM approaches \ours in per-edge accuracy and preserves
more objects, but struggles with front/behind and requires
a 72B planner, a per-video LoRA, and about nine times our
GPU time per clip.
Among the baselines on CogVideoX-5b, LVD falls below the
base model, while Comp-Attn performs close to \ours.
On Wan2.2-A14B, LVD performs best across relations, while
zero-shot \ours is close to Comp-Attn but lowers object count
compared to base (\cref{tab:multirel,tab:transfer}).
However, \ours requires neither layouts nor planners and
transfers to Wan2.2 without retraining the \gmn, while
LVD is sensitive to its background weight
(Suppl. \cref{app:competitors}).

\begin{figure*}[!t]
  \centering
  \includegraphics[width=\linewidth,height=0.90\textheight,keepaspectratio]{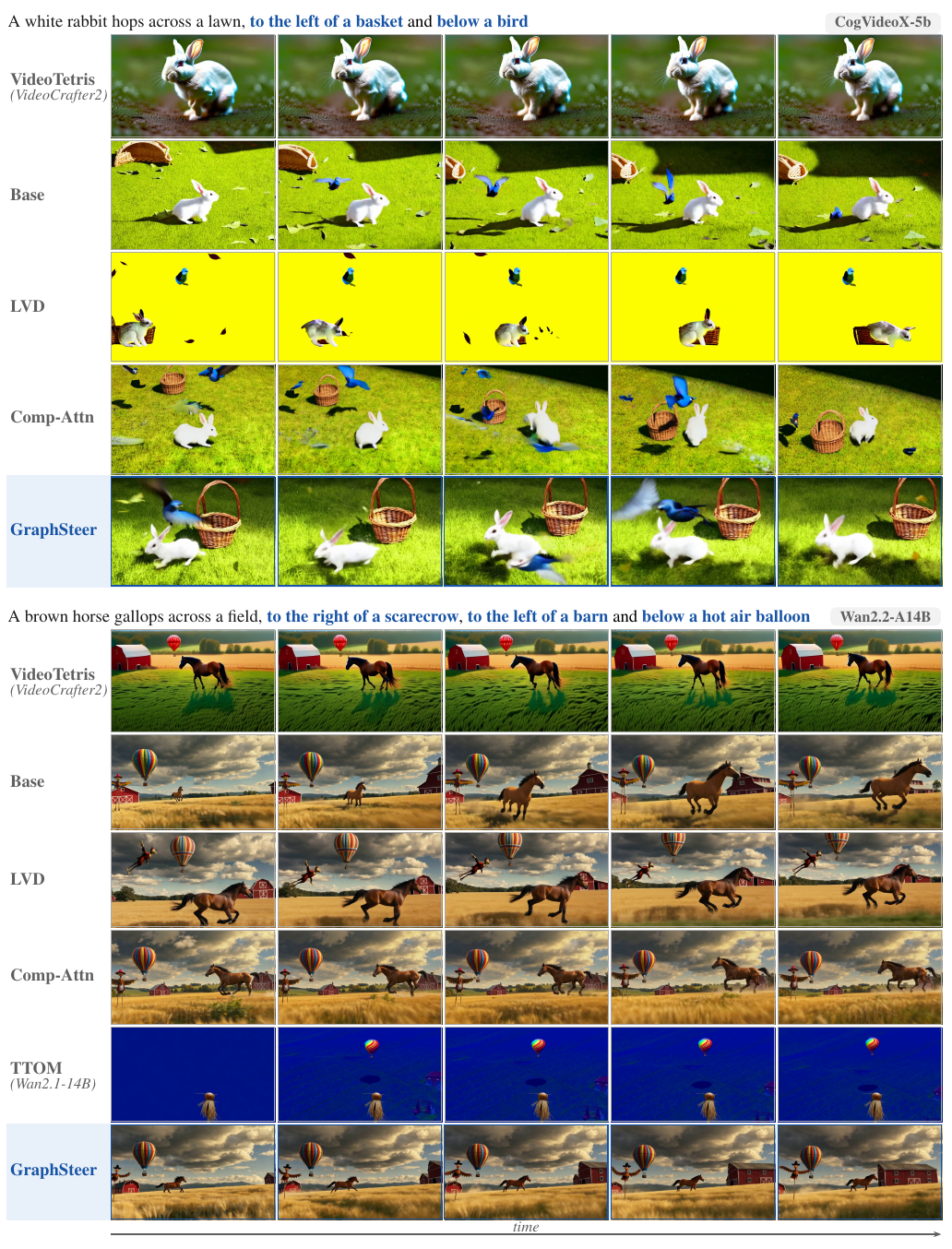}
  \caption{\textbf{Qualitative comparison of all methods on multi-relation prompts}, five frames per clip: CogVideoX-5b (top) and
    Wan2.2-A14B (bottom), each with VideoTetris on VideoCrafter2 in its first row. Clips on one backbone share a seed. TTOM runs on Wan2.1-14B with its own prompt enrichment and seed. LVD at its released settings can remove objects and leave a flat background.
    }
  \label{fig:qual_all}
\end{figure*}

\paragraph{Qualitative results.}
\cref{fig:qual_all} compares all methods on two multi-relation
prompts. On CogVideoX-5b, base and Comp-Attn place the rabbit
to the right of the basket, VideoTetris generates only the
rabbit, and LVD produces a largely uniform yellow background
with severe visual artifacts. \ours is the only method that places
the rabbit to the left of the basket.
On Wan2.2-A14B, base and VideoTetris place the horse to the
right of the barn, while TTOM loses it.
\ours keeps the horse between the scarecrow and the barn,
with the balloon above it. These examples illustrate how layout-based methods can fail
to preserve the requested relations or objects, even when
the planned layout satisfies the prompt.

\subsection{Steering impact on generation quality}
\label{sec:exp_quality}
We evaluate whether steering spatial relations affects
video quality and overall prompt satisfaction.
We use VBench on its standard prompts and our multi-relation
benchmark, together with a VLM-based Global Preference
metric on the multi-relation set.

\begin{table}[t]\centering\small
  \setlength{\tabcolsep}{3pt}
    \begin{tabular*}{\columnwidth}{
  l@{\extracolsep{\fill}}c>{\columncolor{oursrow}}c
}
    \toprule
    & CogVideoX-5b & \quad + \ours \\
    \midrule
    Total score    & 77.63 & \best{78.25} \\
    Quality score  & 81.41 & \best{81.61} \\
    Semantic score & 62.50 & \best{64.78} \\
    \midrule
    Spatial relationship & 39.58 & \best{54.48} \\
    \bottomrule
  \end{tabular*}%
  \caption{\textbf{VBench}. Scores are $\times100$ with the
    official normalization; Supp. \cref{tab:vbench_std_full} lists all 16 dimensions.}
  \label{tab:vbench_std}
\end{table}

\paragraph{VBench.}
Of the $946$ standard VBench prompts \cite{huang2024vbench}
(\cref{tab:vbench_std}), only $288$ state a relation, so most
steered clips equal base. The total score still rises, mainly
through the spatial relationship dimension, with no quality
dimension dropping by more than $0.6$ points.
On the multi-relation benchmark (\cref{tab:vbench}), \ours
achieves the highest CogVideoX-5b quality aggregate, with
higher aesthetic quality, imaging quality, and dynamic degree,
but slightly lower subject and background consistency.
LVD reaches a similar aggregate with less motion, while
Comp-Attn stays close to base.
On Wan2.2-A14B, all methods stay within $0.5$ points of
base. LVD and Comp-Attn score slightly above zero-shot
\ours, mainly through higher imaging quality, but their
clips exhibit less motion.

\begin{figure}[t]
  \centering
  \includegraphics[width=\linewidth]{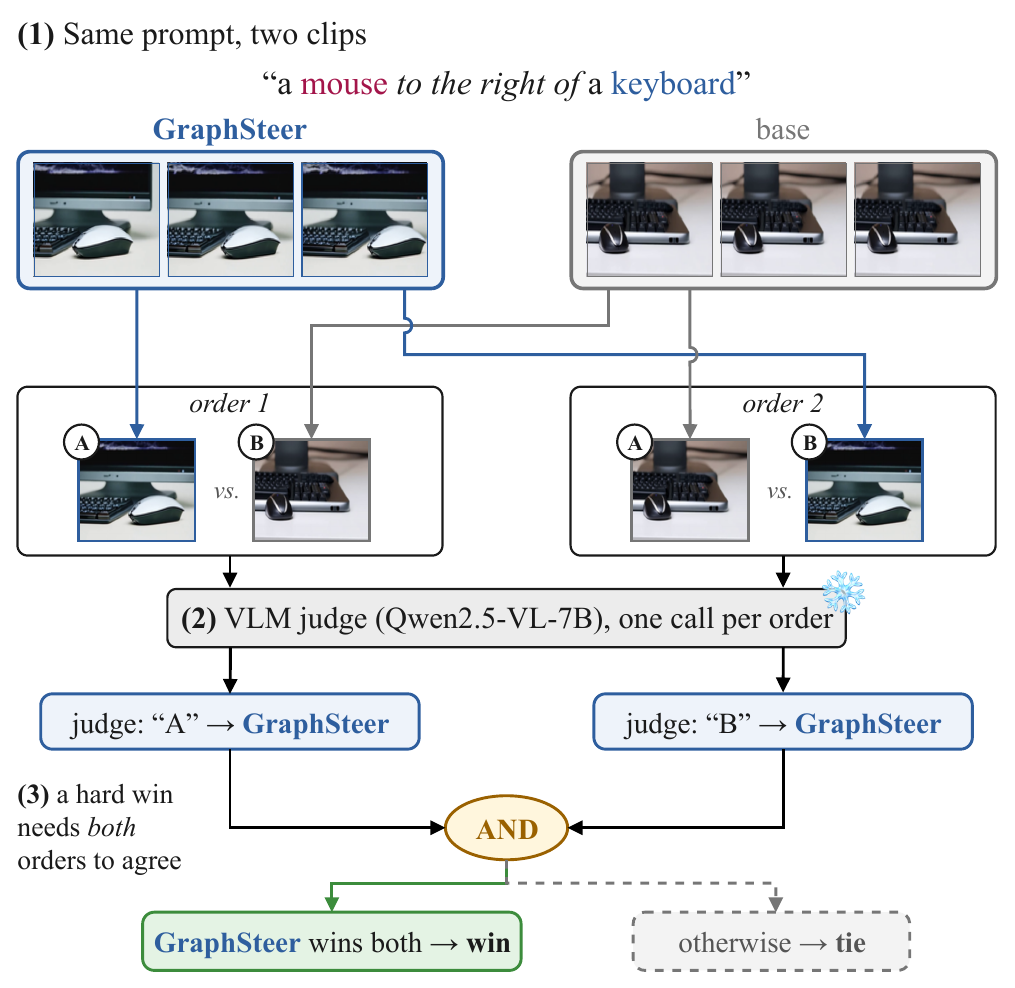}
  \caption{\textbf{VLM-based Global Preference.} 
      A real example where \ours wins in both presentation orders.
    }
  \label{fig:pref_protocol}
\end{figure}

\paragraph{VLM-based Global Preference.}
We introduce a global preference metric based on pairwise
judgments from Qwen2.5-VL-7B, which compares videos
based on relational faithfulness and visual quality.
Since the judge prefers the second video in $70$-$86\%$
of decisive comparisons, we evaluate each pair in both
orders. We report a soft win rate averaging the two
judgments and a hard win rate counting a win only when
both orders agree, assigning a tie otherwise
(\cref{fig:pref_protocol}).
On CogVideoX-5b, the judge prefers \ours over base,
LVD, and Comp-Attn, with the largest margin over LVD
(\cref{tab:preference}).
On Wan2.2-A14B, zero-shot \ours is close to base and LVD
but less preferred than Comp-Attn.
Across backbones, \ours is preferred over VideoTetris,
while TTOM on Wan2.1-14B is preferred over \ours.
TTOM also achieves higher VBench aesthetic quality
(Suppl. \cref{tab:vbench}).

\begin{table}[h]\centering\small
  \setlength{\tabcolsep}{6pt}
      \begin{tabular*}{\columnwidth}{l@{\extracolsep{\fill}}ccc}
    \toprule
    \ours vs. & hard win rate & soft win rate \\
    \midrule
    \multicolumn{3}{l}{\emph{CogVideoX-5b}} \\
    base              & \win{0.746} & \win{0.619} \\
    LVD               & \win{0.863} & \win{0.701} \\
    Comp-Attn         & \win{0.588} & \win{0.557} \\
    \midrule
    \multicolumn{3}{l}{\emph{Wan2.2-A14B} (\ours zero-shot)} \\
    base              & \lose{0.492} & \lose{0.497} \\
    LVD               & \lose{0.475} & \lose{0.488} \\
    Comp-Attn         & \lose{0.442} & \lose{0.483} \\
    \midrule
    \multicolumn{3}{l}{\emph{other backbones}} \\
    VideoTetris (VideoCrafter2) & \win{0.647} & \win{0.574} \\
    TTOM (Wan2.1-14B) & \lose{0.438} & \lose{0.471}        \\
    \bottomrule
  \end{tabular*}%
    \caption{\textbf{VLM-based Global Preference.}
    Hard and soft win rates on the multi-relation benchmark;
    values above $0.5$ favour \ours.
    Within-backbone comparisons share the same base and seed;
cross-backbone comparisons use \ours on CogVideoX-5b.
    }    
  \label{tab:preference}
\end{table}

\subsection{Zero-Shot Transfer Across Backbones}
\label{sec:exp_wan}

Besides Wan2.2-A14B (\cref{tab:competitors,tab:multirel}),
we apply the CogVideoX-trained \gmn to Wan2.1-T2V-14B
\cite{wan2025} and HunyuanVideo \cite{kong2024hunyuanvideo},
changing only the block name (\cref{tab:transfer}).
T2V-CompBench spatial accuracy rises substantially on
HunyuanVideo ($+0.106$) and slightly on Wan2.1.
On multi-relation prompts, the all-relations rate increases
on both backbones, while per-edge accuracy changes little
and object count decreases, indicating that zero-shot transfer
remains partial.
The approach also extends to text-to-image generation:
on SD2.1 and FLUX.1-dev, a \gmn trained for each backbone consistently outperforms Learn-to-Steer and the other baselines on multi-relation prompts (Suppl. \cref{app:images}).

\begin{table}[h]\centering\small
  \setlength{\tabcolsep}{3pt}
  \resizebox{\linewidth}{!}{%
  \begin{tabular}{lccccc}
    \toprule
    & \multicolumn{2}{c}{T2V-CompBench} & \multicolumn{3}{c}{multi-relation} \\
    \cmidrule(lr){2-3}\cmidrule(lr){4-6}
    & Spatial & Numeracy & per-edge & all-rel. & count \\
    \midrule
    Wan2.1-14B                   & 0.569 & 0.502 & 0.188 & 0.017 & \best{0.778} \\
    \oursrow\quad + \ours (zero-shot)    & \best{0.595} & \best{0.512} & \best{0.212} & \best{0.050} & 0.764 \\
    \midrule
    HunyuanVideo                 & 0.610 & \best{0.313} & 0.350 & 0.058 & \best{0.838} \\
    \oursrow\quad + \ours (zero-shot)    & \best{0.715} & 0.281 & \best{0.358} & \best{0.117} & 0.754 \\
    \bottomrule
  \end{tabular}}
  \caption{\textbf{Zero-shot transfer} with identical guidance settings (T2V-CompBench $n{=}200$,
    multi-relation $n{=}120$).}
  \label{tab:transfer}
\end{table}

%% file: sec/5_conclusion.tex
\section{Conclusion}
\label{sec:conclusion}
We introduced \ours, a layout-free method
for jointly steering spatial relations in frozen video
diffusion models. 
By constructing a scene graph from the prompt and using
a learned \gmn, our method scores multiple relations
directly from cross-attention and guides generation
through a single joint loss.
Our experiments show that \ours
improves relational accuracy on CogVideoX-5b, outperforming
layout-based methods without requiring predefined layouts.
Crucially, the same \gmn transfers zero-shot to different
video backbones, including Wan2.2, Wan2.1, and HunyuanVideo,
without any retraining, although improvements vary
across models.

%% file: sec/X_suppl.tex
\renewcommand{\thefigure}{A\arabic{figure}}
\renewcommand{\theHfigure}{A\arabic{figure}}
\renewcommand{\thesection}{A\arabic{section}}
\renewcommand{\theHsection}{A\arabic{section}}
\renewcommand{\theequation}{A\arabic{equation}}
\renewcommand{\theHequation}{A\arabic{equation}}
\renewcommand{\thetable}{A\arabic{table}}
\renewcommand{\theHtable}{A\arabic{table}}

\setcounter{equation}{0}
\setcounter{figure}{0}
\setcounter{table}{0}
\setcounter{section}{0}
\ifdefined\suppinmain\else\setcounter{page}{1}\fi %

\maketitlesupplementary
\appendix
\renewcommand{\thesection}{A\arabic{section}}
\renewcommand{\theHsection}{A\arabic{section}}

\newcommand{\fixh}[1]{\texorpdfstring{#1}{#1}}
\renewcommand{\topfraction}{0.9}
\renewcommand{\dbltopfraction}{0.9}
\renewcommand{\textfraction}{0.1}
\renewcommand{\floatpagefraction}{0.8}
\renewcommand{\dblfloatpagefraction}{0.8}

This document contains the details and further experiments that did not fit in the main paper. \cref{app:details} contains the training data, including the synthetic clips, the \gmn architecture and the relation parser with its prompts. \cref{app:ablations} ablates the components of \ours and \cref{app:results} contains further results: the remaining T2V-CompBench and VBench dimensions, the multi-relation benchmark, the human preference study, video examples and transfer to image generation. \cref{app:baselines} describes how we ran each of the competitors and compares their computational cost. \cref{app:hyperparamters} gives the generation settings of each video backbone and the block the \gmn reads on each of them, the hyperparameters used to train the \gmn and to guide generation, and a comparison of four \gmn designs. \cref{app:limits} contains the limitations and societal impact. The code and \gmn checkpoints will be made available upon acceptance of the paper.

\section{Implementation details}
\label{app:details}
\subsection{Training data}
\label{app:data}
The \gmn is trained on real videos from VidOR and ImageNet-VidVRD \cite{shang2019vidor,shang2017vidvrd}, plus about $600$ synthetic clips labelled from GroundingDINO boxes (\cref{sec:experiments}). The VidOR statistics in \cref{tab:data_stats} shaped two design choices. First, a single edge is often labelled on several axes at once, for example on depth and proximity, so we predict each axis with its own head instead of one softmax over all relations. Second, about a quarter of the depth labels change sign within a video, so we split each video into windows and keep a label only when it holds in at least $70\%$ of the frames of its window.

\begin{table}[h]\centering\small
  \setlength{\tabcolsep}{4pt}
  \begin{tabular}{lr}
    \toprule
    statistic (VidOR, $1{,}663$ windowed scenes) & value \\
    \midrule
    edges labelled on depth \emph{and} proximity & $3{,}250$ \\
    edges labelled on depth \emph{and} motion & $2{,}460$ \\
    edges labelled on depth, motion \emph{and} proximity & $1{,}098$ \\
    depth edges whose sign flips within a video & $26\%$ \\
    \bottomrule
  \end{tabular}
  \caption{Statistics of the real clips from VidOR.}
  \label{tab:data_stats}
\end{table}
\paragraph{Synthetic clips.} Left/right is by far the rarest axis in the real data: we have just 1312 horizontal edges from 165 VidOR and VidVRD clips. We generate synthetic two-entity clips to fill this in. The prompt is of the form ``A to the left of B" or ``A to the right of B" with A and B being drawn from a set of 26 common nouns (dog, car, bench, umbrella, laptop, etc.) which CogVideoX-5b can render and GroundingDINO can detect. CogVideoX-5b produces 990 clips (600 left, 390 right). We do not trust the prompt as a label. We use GroundingDINO (Swin-T) to detect both entities on 8 frames sampled uniformly across the clip, looking at the horizontal centre of the two highest scoring boxes. A frame only votes if the centers are at least 5\% of the frame width apart. A clip is retained if at least one frame votes and at least 60\% of the votes agree. The label of the clip is the majority side. This gives us 634 clips (64\%). 354 are discarded as one of the entities is missing or the two are too close and 2 are discarded because the votes disagree. In 255 of the clips that we keep (40\%) the rendered side is the opposite of the prompt so for 40\% of the synthetic data the prompt would have been wrong. The clips that we keep take the number of scenes with a horizontal edge from 165 to 799.

For the image experiments (\cref{app:images}), we use GQA \cite{hudson2019gqa} scenes inverted through SD2.1, as in Learn-to-Steer \cite{learn2steer}, plus FLUX.1-schnell scenes labelled from detected layouts.

\subsection{\gmn architecture}
\label{app:reader}
The \gmn reads the attention of one transformer block. On CogVideoX-5b we use block $15$ of $42$ with all $48$ heads, the block whose attention best matched detected boxes in a probe, and we keep the same index on Wan2.2-A14B without re-tuning. To check this choice, we ran the probe on all $40$ cross-attention blocks of Wan2.2-A14B over $60$ VidVRD test scenes. For each pair of entities of different categories, we compare the order of their attention centroids (computed over the top $5\%$ cells) with the order of their ground-truth boxes. Early blocks ($0$--$7$) agree more often on the horizontal axis than block $15$ ($0.65$ against $0.51$, with chance at $0.50$), but each block has only $27$ left/right pairs, so the uncertainty is about $\pm0.09$. On the vertical axis the two blocks agree equally often. Guiding with block $7$ does not help either: the T2V-CompBench spatial gain is not significant, and per-edge accuracy on our multi-relation benchmark drops slightly. We therefore keep block $15$.

\paragraph{Entity encoder.} Each map is resampled to $8{\times}16{\times}16$, normalized by its mass, and projected to $d{=}256$. A two-layer transformer without head identity encodes the maps of all heads, and a learned query pools them, so the encoder accepts any number of heads. During training we simulate other backbones by sampling subsets of heads, changing the grid, zooming, and making the attention sharper or flatter. On CogVideoX this encoder steers at least as well as one trained on the native grid ($0.784$ against $0.738$ on $60$ left/right prompts).

\paragraph{Edge features.} Each edge is scored from the two node embeddings, their difference, the mean embedding over all nodes, and $16$ statistics of the two attention footprints. The graph neural network has $2$ graph-attention layers with $4$ heads.

\subsection{Parsing prompt to graph}
\label{app:parser}
Qwen2.5-VL-7B, on text only and with greedy decoding, returns the subject, object, and
(axis, sign) target of every relation as JSON. Its instructions include rules for shared subjects and for support relations such as ``on'' (\cref{app:prompts}).

\paragraph{Agreement with gold graphs.} Given gold entities, a parse counts as correct only when the whole graph matches the gold one (\cref{tab:parser}). An edge stated in the reverse direction, such as ``$B$ right of $A$'' for ``$A$ left of $B$'', counts as a match.

\begin{table}[t]\centering\small
  \setlength{\tabcolsep}{4pt}
  \resizebox{\linewidth}{!}{%
  \begin{tabular}{lcc}
    \toprule
    parser & multi-relation & T2V-CompBench spatial \\
    \midrule
    rule-based & $0.0\%$ ($0/120$)   & $\mathbf{100\%}$ ($200/200$) \\
    Qwen2.5-3B-Instruct    & $95.8\%$ ($115/120$) & $87.0\%$ ($174/200$) \\
    \oursrow Qwen2.5-VL-7B-Instruct & $\mathbf{99.2\%}$ ($119/120$) & $97.5\%$ ($195/200$) \\
    \bottomrule
  \end{tabular}}
  \caption{Exact-graph agreement of relation parsers with gold graphs. The shaded row is the parser used by \ours.}
  \label{tab:parser}
\end{table}

\noindent The rule-based parser is a simple baseline without a language model: it finds the relation word in the prompt and links it to the nouns closest to it on either side. It is exact on T2V-CompBench, where each prompt states one relation, but on every multi-relation prompt it attaches at least one relation to the wrong noun. The VLM parser handles both. CogVideoX runs use the Qwen2.5-VL-7B parser. The parser also maps $19$ of $20$ held-out phrasings, such as ``atop'' and ``occluded by'', to the right target, where chance is $0.11$ (one of nine axis and sign targets).

\subsection{Prompts}
\label{app:prompts}
These are the instructions sent to each language model. All models run
locally.

\paragraph{Entity extraction} (Qwen2.5-3B-Instruct).
\begin{quote}\footnotesize\itshape
You are a parser for compositional text-to-video generation. Given a prompt, output a JSON
object with the list of entities: one item per object, with an id (\texttt{noun\_n}), the noun
phrase exactly as it appears in the prompt, and whether it is a background region (sky, ground,
room, street, \ldots) or an object. Bind every adjective and verb to the entity it modifies.
Output JSON only.
\end{quote}

\paragraph{Relation parsing} (Qwen2.5-VL-7B-Instruct).
\begin{quote}\footnotesize\itshape
You extract spatial relations from a text-to-video prompt. You are given the prompt and its
entities. For every stated relation output its subject, object, the relation words, and an
axis and pole from this taxonomy: horizontal (left, right), vertical (above, below), depth
(in front of, behind), motion (toward, away), proximity (next to). Rules: the subject is the
entity the relation is about; when one subject has several relations joined by ``and'', all of
them have that subject, not the nearest noun. Emit one item per stated relation and never invent
one. ``On'', ``on top of'', ``atop'' mean vertical/above; ``under'', ``beneath'' mean
vertical/below; use proximity only for ``next to'', ``beside'', ``near''. Omit relations that
fit no axis, such as holding or riding. Output JSON only.
\end{quote}

\paragraph{Preference judge} (Qwen2.5-VL-7B-Instruct).
\begin{quote}\footnotesize\itshape
Two AI systems generated videos (shown as frame strips) for the same text prompt.
Text prompt: ``\textup{$\langle$prompt$\rangle$}''. Compare video A and video B on both of these,
weighted equally: (1) faithfulness: are all the mentioned objects present, and are their spatial
relations as described? (2) overall quality: natural composition, no distorted or merged
objects, no artifacts. Which one is better overall? Reply with exactly one word: A, B, or TIE.
\end{quote}

\section{Ablations}
\label{app:ablations}
We ablate three components of \ours: the graph layers of the \gmn (\cref{app:between}), the presence term (\cref{app:presence}), and the number of guided steps (\cref{app:steps}). Each study changes one component and keeps the rest of the pipeline fixed.

\subsection{Message passing}
\label{app:between}
In most of our prompts, a relation is decided by the two entities of its edge, so a \gmn that scores each pair on its own already sees everything it needs. \emph{Between} is different. ``$S$ between $B$ and $C$'' becomes two edges, $S{\to}B$ and $S{\to}C$, with the same label, and whether $S{\to}B$ means \emph{between} or \emph{right of} depends on $C$. We use this relation to test what the graph layers add.

We train two SD2.1 scorers that differ only in the number of graph-attention layers: $L{=}2$, as in \ours, and $L{=}0$, where each edge is scored from its two node embeddings alone (per-pair). Both learn five labels (left of, right of, above, beneath, between) from the same $3{,}135$ scenes with a class-balanced loss; \emph{between} makes up $8.6\%$ of the edges. A scene that exists under both labelings is kept once, so identical attention maps never appear in both training and validation.

\paragraph{Reading.} Without graph layers, the scorer almost never predicts \emph{between}: its validation recall on this class is $0.014$, against $0.608$ for the graph scorer. Whether an edge means \emph{between} or \emph{left of} depends on the third entity, which only the graph scorer sees. Balanced accuracy over the five classes is about the same for both scorers ($0.526$ and $0.511$), so we make no claim about reading accuracy in general and test what the graph layers change when steering.

\paragraph{Steering.} We steer SD2.1 with each scorer on $60$ prompts of the form ``$A$ between $B$ and $C$'', with a random axis and one seed. An image counts as correct when the subject lies strictly inside the interval spanned by the two references on the axis that separates them. Degenerate layouts are skipped, and the two edges count as one constraint. We then add two distractor entities that no relation constrains (\cref{tab:between_steer}).

\begin{table}[t]\centering\small
  \setlength{\tabcolsep}{4pt}
  \begin{tabular}{lcc}
    \toprule
    \emph{between} satisfied & 3 entities & 5 entities \\
    \midrule
    base SD2.1             & 0.250 & 0.150 \\
    per-pair ($L{=}0$)     & 0.267 & 0.150 \\
    \oursrow graph ($L{=}2$)        & \best{0.550} & \best{0.550} \\
    \bottomrule
  \end{tabular}
  \caption{ Message passing enables reading ternary relations.}
  \label{tab:between_steer}
\end{table}

The per-pair scorer stays at the base rate, while the graph scorer more than doubles it. The per-pair signal is not weak: it pushes the latent toward left or right, the classes it can represent, but that does not put the subject between the references. With two distractors, base and per-pair accuracy drop, while the graph scorer keeps its accuracy. The gap between the two scene sizes is not significant at $n{=}60$, so we only claim that the advantage of the graph scorer holds at both sizes. These are single-seed image experiments with an SD2.1 scorer, not the video scorer used in the rest of the paper.

\subsection{Presence term}
\label{app:presence}
The \gmn sees maps divided by their own mass, so an entity whose attention has almost vanished can still satisfy a relation. \cref{tab:presence} removes the presence term of \cref{eq:pres} ($\lambda{=}0$) and keeps everything else fixed (same \gmn, guided steps, step size and seed) on CogVideoX-5b. Steering without the term already raises accuracy on both benchmarks, and the term raises it further.

\begin{table}[t]\centering\small
  \setlength{\tabcolsep}{4pt}
  \resizebox{\linewidth}{!}{%
  \begin{tabular}{lccc}
    \toprule
    & per-edge$\uparrow$ & all-rel.$\uparrow$ & T2V-CompBench spatial$\uparrow$ \\
    \midrule
    base                    & 0.225 & 0.058 & 0.545 \\
    $\lambda{=}0$           & 0.366 & 0.143 & 0.590 \\
    \oursrow $\lambda{=}4$ (ours)    & \best{0.392} & \best{0.150} & \best{0.631} \\
    \bottomrule
  \end{tabular}}
  \caption{\textbf{Presence term.} CogVideoX-5b, one seed. The first two columns are on our multi-relation benchmark: per-edge accuracy and all-relations rate. The last is T2V-CompBench spatial.}
  \label{tab:presence}
\end{table}

The threshold $0.5$ in \cref{eq:pres} also matters. With a threshold of $1$, every drop in attention mass is penalized, including the drops that steering needs to move an entity, and left/right accuracy on a $60$-prompt T2V-CompBench subset falls from $0.729$ to $0.630$.

\subsection{Number of guided steps}
\label{app:steps}
\ours updates the latent during the first $8$ of $50$
denoising steps. We vary the number of guided steps,
keeping the \gmn, presence term, step size, and seed fixed
(\cref{tab:steps}). On the multi-relation benchmark with
CogVideoX-5b, four steps already provide most of the
improvement over base, while $16$ steps achieve higher
per-edge accuracy. However, neither differs significantly
from $8$ steps. On a $60$-prompt subset of T2V-CompBench spatial, $4$ and $8$ steps score the same, but $16$ steps are significantly better than $8$. A longer window can therefore help on CogVideoX, and we do not claim that $8$ steps is optimal. On Wan2.2-A14B, the $16$-step window covers every step of the high-noise expert, and it does not help: $4$, $8$ and $16$ steps score $0.672$, $0.675$ and $0.663$, all within noise of each other and of base at $n{=}60$. Since each additional guided step requires an extra forward and backward pass, we use $8$ steps to balance accuracy and computational cost. Each guided step adds about $0.14$ minutes per clip on CogVideoX and $0.25$ on Wan2.2, so $8$ steps raise generation time by $26\%$ and $14\%$ over base, $16$ steps by $52\%$ and $28\%$, and $4$ steps by $17\%$ and $9\%$.

\begin{table}[t]\centering\small
  \setlength{\tabcolsep}{4pt}
  \resizebox{\linewidth}{!}{%
  \begin{tabular}{lcccccc}
    \toprule
    & \multicolumn{2}{c}{CogVideoX-5b, multi-rel.} & \multicolumn{2}{c}{CogVideoX-5b} & \multicolumn{2}{c}{Wan2.2-A14B} \\
    \cmidrule(lr){2-3}\cmidrule(lr){4-5}\cmidrule(lr){6-7}
    guided steps & per-edge$\uparrow$ & all-rel.$\uparrow$ & spatial$\uparrow$ & min/clip$\downarrow$ & spatial$\uparrow$ & min/clip$\downarrow$ \\
    \midrule
    base (0)                 & 0.225 & 0.058 & 0.547 & 4.2 & 0.653 & 14.9 \\
    $0$--$3$                 & 0.375 & 0.133 & 0.637 & 4.9 & 0.672 & 16.3 \\
    \oursrow $0$--$7$ (ours) & 0.392 & 0.150 & 0.636 & 5.3 & \best{0.675} & 17.0 \\
    $0$--$15$                & \best{0.467} & \best{0.175} & \best{0.689} & 6.4 & 0.663 & 19.0 \\
    \bottomrule
  \end{tabular}}
  \caption{\textbf{Number of guided steps.} One seed. The first two columns are on our multi-relation benchmark ($120$ prompts). The spatial columns of the other two backbones are T2V-CompBench spatial on the same $60$ prompts. The min/clip columns values are computed as the mean generation time per clip on one A100 GPU, model loading excluded.}
  \label{tab:steps}
\end{table}

\section{Additional results}
\label{app:results}
\subsection{T2V-CompBench and VBench}
\label{app:compbench7}
\cref{tab:compbench_7dim} reports all seven T2V-CompBench dimensions. The gain is concentrated in spatial and numeracy, and the other five dimensions stay within $0.02$ of base.

\begin{table*}[t]\centering\small
  \setlength{\tabcolsep}{7pt}
  \begin{tabular}{lccccccc|c}
    \toprule
    & Con-attr & Dyn-attr & \textbf{Spatial} & Motion & Action & Interact & Numeracy & \textbf{Avg} \\
    \midrule
    CogVideoX-5b & 0.694 & 0.013 & 0.545 & 0.243 & 0.517 & 0.569 & 0.278 & 0.408 \\
    \oursrow\quad + \ours & \best{0.714} & \best{0.014} & \best{0.631} & \best{0.253} & \best{0.528} & \best{0.571} & \best{0.346} & \best{0.432} \\
    \bottomrule
  \end{tabular}
  \caption{All seven T2V-CompBench dimensions with the official evaluators and the same seed. The gain is in spatial and numeracy.}
  \label{tab:compbench_7dim}
\end{table*}

On the $946$ standard VBench prompts (\cref{tab:vbench_std_full}), only spatial relationship moves substantially. Human action also rises, and every other dimension changes by less than $1.5$ points.

\begin{table}[t]\centering\small
  \setlength{\tabcolsep}{6pt}
  \begin{tabular}{lc>{\columncolor{oursrow}}c}
    \toprule
    dimension & CogVideoX-5b & + \ours \\
    \midrule
    \multicolumn{3}{l}{\emph{quality}} \\
    \quad subject consistency & 95.19 & \best{95.48} \\
    \quad background consistency & 95.75 & 95.75 \\
    \quad temporal flickering & \best{97.62} & 97.60 \\
    \quad motion smoothness & \best{97.74} & 97.71 \\
    \quad dynamic degree & 76.39 & \best{77.78} \\
    \quad aesthetic quality & \best{57.93} & 57.34 \\
    \quad imaging quality & 57.77 & \best{58.77} \\
    \midrule
    \multicolumn{3}{l}{\emph{semantic}} \\
    \quad object class & 70.97 & 70.97 \\
    \quad multiple objects & 43.06 & 43.06 \\
    \quad human action & 65.00 & \best{69.00} \\
    \quad color & 87.65 & 87.65 \\
    \quad spatial relationship & 39.58 & \best{54.48} \\
    \quad scene & 34.08 & 34.08 \\
    \quad appearance style & 23.41 & \best{23.44} \\
    \quad temporal style & 23.34 & 23.34 \\
    \quad overall consistency & 25.03 & \best{25.57} \\
    \bottomrule
  \end{tabular}
  \caption{All 16 standard VBench dimensions on CogVideoX-5b. Only spatial relationship moves substantially. Color skips clips in which the object is not detected.}
  \label{tab:vbench_std_full}
\end{table}

\subsection{Depth relations}
\label{app:depth}
Since each axis has its own head (\cref{app:data}), the \gmn can read depth from attention: on front/behind pairs, the depth head reaches a balanced accuracy of $0.705$, while a head that only knows the horizontal and vertical axes, as in image methods \cite{learn2steer}, labels $87\%$ of these pairs as vertical. Reading depth, however, does not mean we can steer it. On the multi-relation benchmark with CogVideoX-5b (\cref{tab:multirel}), the gain over base is largest on left/right ($+0.242$), smaller on above/below ($+0.136$) and small on front/behind ($+0.044$). We therefore claim that depth is readable from attention, not that it is steerable.

\subsection{Multi-relation benchmark}
\label{app:multirel}
The gold graph of each prompt is used only for scoring. On $8$ frames per clip, GroundingDINO boxes decide left/right and above/below. Front/behind follows T2V-CompBench: Depth-Anything inside SAM masks, scored only when the two boxes overlap. A relation holds when it is satisfied in at least half of the frames in which both objects appear. With arbitrary object pairs, as in \cite{learn2steer}, the base video model cannot compose the scene and every method scores zero, so each prompt describes a plausible natural scene.

Under the official rule, a front/behind relation fails when the two boxes do not overlap, even if the depth order is right. \cref{tab:multirel_order} also scores these edges by depth order alone. This only changes front/behind edges and raises every method's scores, and \ours stays first under both rules.

\begin{table}[t]\centering\small
  \setlength{\tabcolsep}{4pt}
  \resizebox{\linewidth}{!}{%
  \begin{tabular}{lcccc}
    \toprule
    & \multicolumn{2}{c}{official (overlap)} & \multicolumn{2}{c}{order only} \\
    \cmidrule(lr){2-3}\cmidrule(lr){4-5}
    method & per-edge & all-rel. & per-edge & all-rel. \\
    \midrule
    VideoTetris (VideoCrafter2) & 0.307 & 0.101 & 0.332 & 0.109 \\
    TTOM (Wan2.1-14B)           & 0.375 & 0.100 & 0.425 & 0.158 \\
    \midrule
    CogVideoX-5b        & 0.225 & 0.058 & 0.250 & 0.092 \\
    \quad + LVD$^\ast$       & 0.192 & 0.058 & 0.200 & 0.075 \\
    \quad + Comp-Attn$^\ast$ & 0.379 & 0.133 & 0.400 & 0.158 \\
    \oursrow\quad + \ours       & \best{0.392} & \best{0.150} & \best{0.450} & \best{0.192} \\
    \bottomrule
  \end{tabular}}
  \caption{Scores with and without the overlap requirement for front/behind. Rows as in \cref{tab:multirel}; bold marks the best value in each column. \ours is first under both rules.}
  \label{tab:multirel_order}
\end{table}

\begin{figure*}[t]
  \centering
  \includegraphics[width=\linewidth]{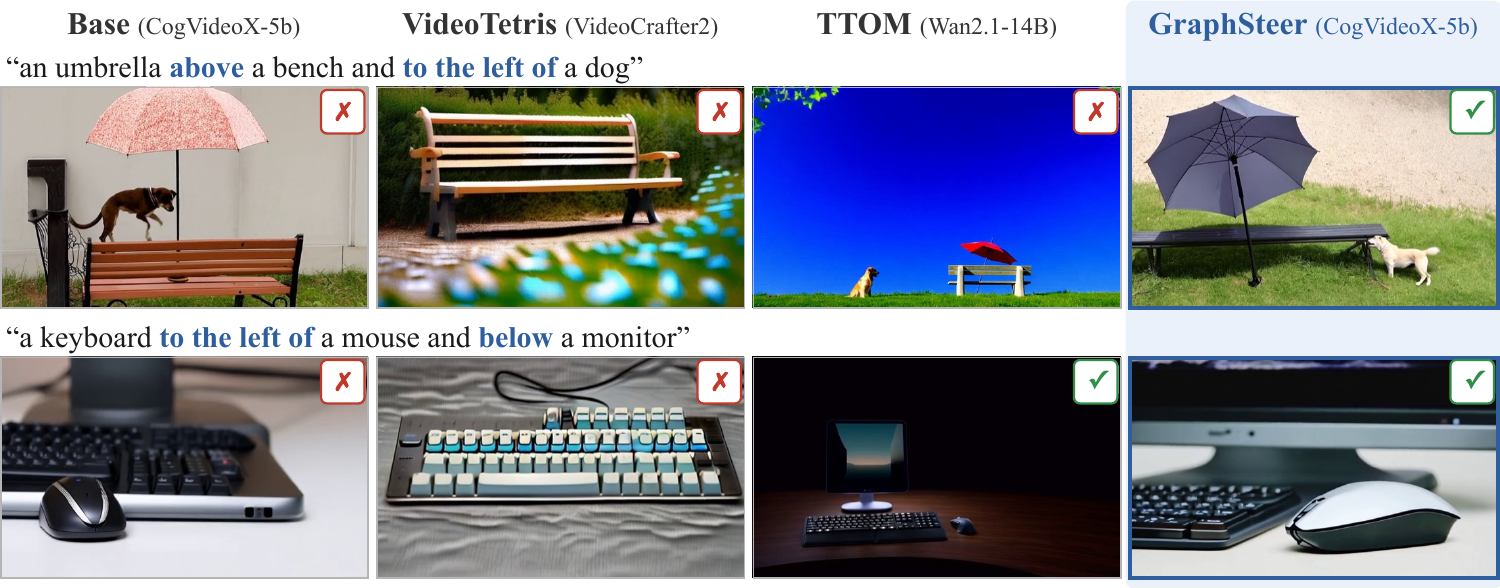}
  \caption{\textbf{Multi-relation prompts, all methods.} Middle frame of each clip; marks show the
    detector's verdict on each relation.}
  \label{fig:qual_comp}
\end{figure*}

\cref{fig:qual_comp} shows the detector's verdicts on two multi-relation prompts. \ours is marked correct on both, TTOM only on the second, and base and VideoTetris on neither. On the first prompt, VideoTetris and TTOM also drop or shrink one of the objects.

\cref{tab:vbench} reports the VBench quality dimensions of the clips in \cref{tab:multirel}.

\begin{table}[t]\centering\footnotesize
  \setlength{\tabcolsep}{2pt}
  \resizebox{\linewidth}{!}{%
  \begin{tabular}{lccccccc|c}
    \toprule
    & \multicolumn{2}{c}{frame-wise} & \multicolumn{5}{c|}{temporal} & \\
    \cmidrule(lr){2-3}\cmidrule(lr){4-8}
    & aesth. & imag. & dyn. & subj. & bg. & smooth. & flick. & \textbf{Quality} \\
    \midrule
    VideoTetris          & 0.511 & 0.625 & 0.605 & 0.952 & 0.962 & 0.955 & 0.937 & 77.11 \\
    TTOM                 & 0.585 & 0.635 & 0.008 & 0.965 & 0.968 & 0.995 & 0.995 & 78.73 \\
    \midrule
    CogVideoX-5b         & 0.438 & 0.591 & 0.108 & 0.970 & 0.975 & 0.993 & 0.989 & 76.44 \\
    \quad + LVD$^\ast$       & 0.521 & 0.652 & 0.183 & 0.972 & 0.973 & 0.992 & 0.990 & 79.24 \\
    \quad + Comp-Attn$^\ast$ & 0.466 & 0.611 & 0.100 & 0.968 & 0.971 & 0.993 & 0.990 & 77.04 \\
    \oursrow\quad + \ours        & 0.509 & 0.662 & 0.292 & 0.962 & 0.960 & 0.993 & 0.983 & \best{79.32} \\
    \midrule
    Wan2.2-A14B          & 0.537 & 0.660 & 0.050 & 0.992 & 0.981 & 0.997 & 0.998 & 79.70 \\
    \quad + LVD$^\ast$       & 0.538 & 0.696 & 0.025 & 0.993 & 0.983 & 0.997 & 0.998 & \best{80.15} \\
    \quad + Comp-Attn$^\ast$ & 0.538 & 0.696 & 0.017 & 0.993 & 0.982 & 0.997 & 0.998 & 80.03 \\
    \oursrow\quad + \ours (zero-shot) & 0.542 & 0.668 & 0.050 & 0.991 & 0.981 & 0.997 & 0.998 & 79.86 \\
    \bottomrule
  \end{tabular}}
    \caption{\textbf{Video quality on multi-relation prompts.}
    VBench quality dimensions for the clips in
    \cref{tab:multirel}. Dyn.\ measures motion; Quality is
    VBench's weighted aggregate ($\times100$).
    Bold marks the best within each backbone.
    $^\ast$Our runs on the same base and seed.}
  \label{tab:vbench}
\end{table}

\subsection{Human preference study}
\label{app:human}
\begin{figure*}[t]
  \centering
  \begin{minipage}[b]{0.30\linewidth}\centering
    \includegraphics[width=\linewidth]{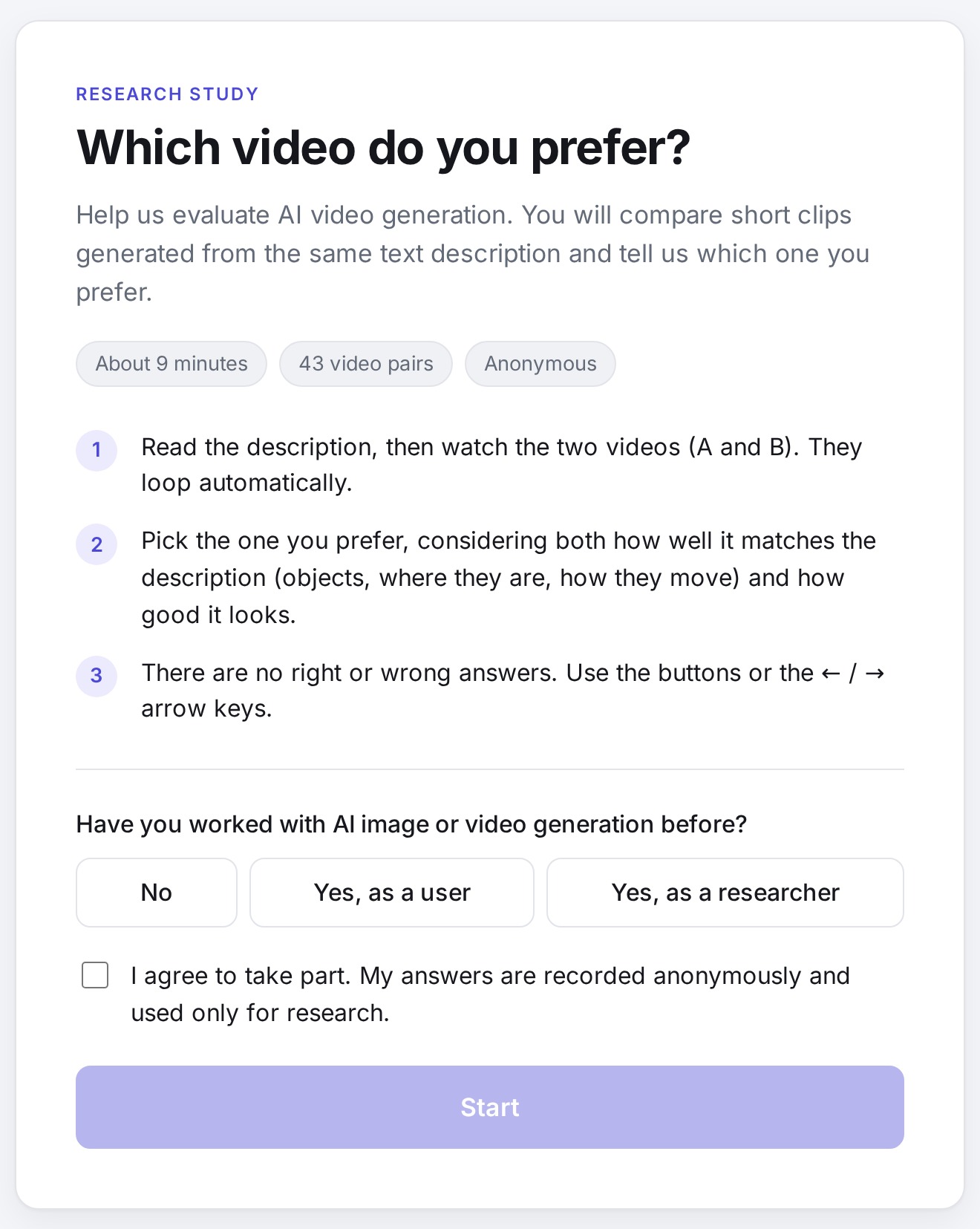}\\[2pt]
    {\small (a) start screen}
  \end{minipage}\hfill
  \begin{minipage}[b]{0.675\linewidth}\centering
    \includegraphics[width=\linewidth]{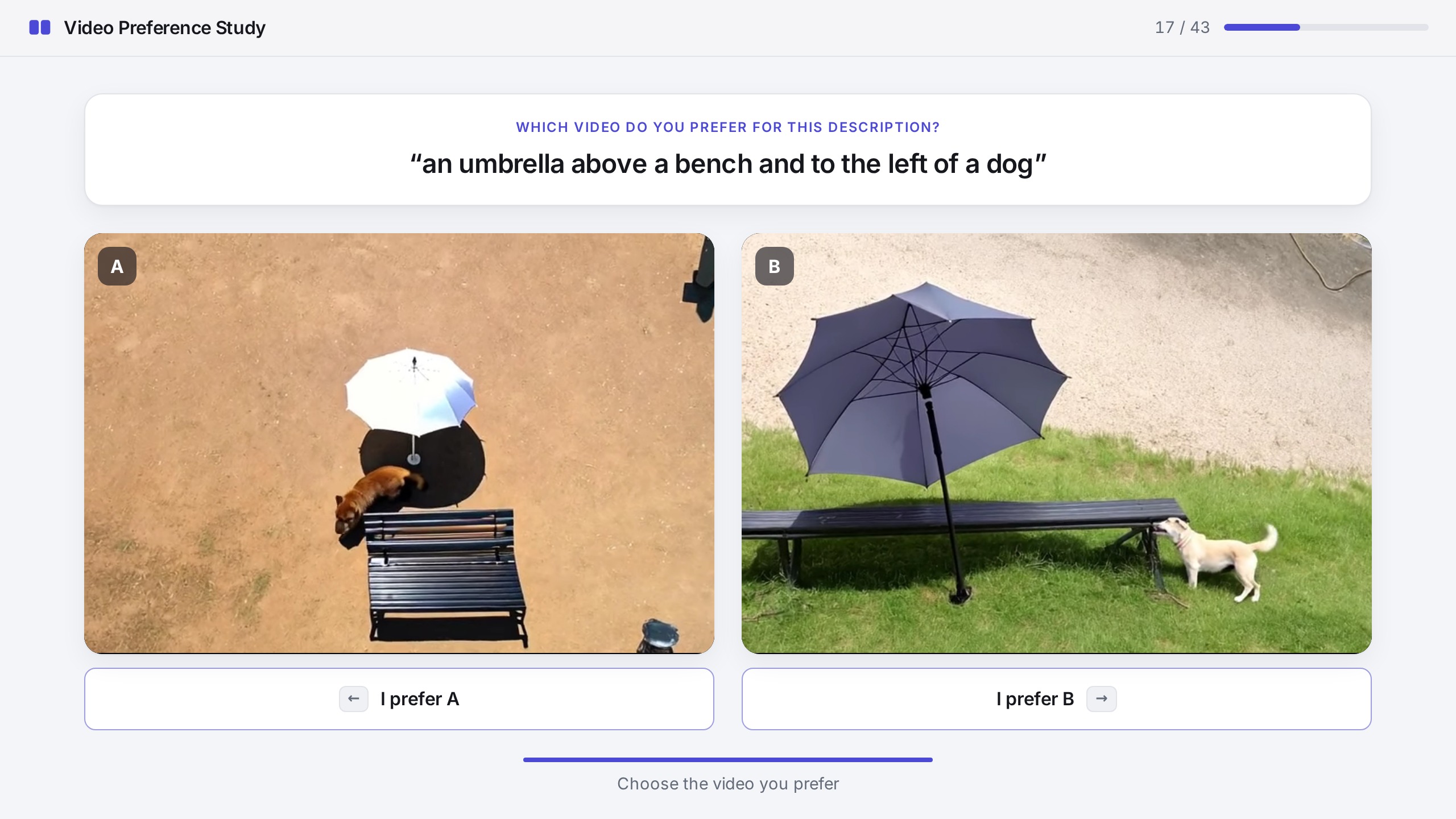}\\[2pt]
    {\small (b) one trial}
  \end{minipage}
  \caption{\textbf{Human preference study interface.} (a) Start screen with the instructions shown to participants. (b) One trial: base (A) vs.\ \ours (B) on CogVideoX-5b. Participants do not know which method made which clip, and the sides are random.}
  \label{fig:human_study}
\end{figure*}

To check whether the VLM judge of \cref{tab:preference} agrees with people, we ran a pairwise study on the same clips. \ours on CogVideoX-5b is compared with five methods: base, LVD and Comp-Attn on CogVideoX-5b, VideoTetris on VideoCrafter2, and TTOM on Wan2.1-14B. The Wan2.2-A14B rows of \cref{tab:preference} are not part of the study. Each trial shows two clips generated from the same multi-relation prompt, together with the prompt, and the participant picks the one they prefer (\cref{fig:human_study}). The instructions ask participants to consider both how well a clip matches the description (the objects, where they are, how they move) and how good it looks, the two criteria the VLM judge also uses.

Each participant sees $40$ pairs, $8$ per baseline, each from a different prompt, with the two clips placed on random sides. Three pairs from the first half are shown again in the second half with the sides swapped. We use these repeats only to measure each participant's consistency and do not count them in the rates. Clip file names are hashed, so the page does not reveal the method, and the choice buttons unlock only after both clips have played for $3.5$\,s. VideoTetris clips play at $10$\,fps, the default of its released code; the frames are the ones scored in the main paper. Before starting, participants say whether they have used AI image or video generation before.

\cref{tab:human} gives, for each baseline, the fraction of answers that prefer \ours. Participants cannot declare a tie. The proposed VLM-based assessment demonstrates close alignment with human results; specifically, the scores are highly similar when evaluated with the same backbone. Furthermore, the findings indicate that human judgments are more decisive, as their rates deviate further from $0.5$.

\begin{table}[t]\centering\small
  \setlength{\tabcolsep}{4pt}
  \resizebox{\linewidth}{!}{%
  \begin{tabular}{lccc}
    \toprule
    & \multicolumn{2}{c}{VLM (\cref{tab:preference})} & human \\
    \cmidrule(lr){2-3}\cmidrule(lr){4-4}
    \ours vs. & hard & soft & preference \\
    \midrule
    \multicolumn{4}{l}{\emph{CogVideoX-5b}} \\
    base                        & \win{0.746} & \win{0.619} & \win{0.795} \\
    LVD                         & \win{0.863} & \win{0.701} & \win{0.962} \\
    Comp-Attn                   & \win{0.588} & \win{0.557} & \win{0.614} \\
    \midrule
    \multicolumn{4}{l}{\emph{other backbones}} \\
    VideoTetris (VideoCrafter2) & \win{0.647} & \win{0.574} & \win{0.811} \\
    TTOM (Wan2.1-14B)           & \lose{0.438} & \lose{0.471} & \lose{0.304} \\
    \bottomrule
  \end{tabular}}
  \caption{\textbf{Human vs.\ VLM preference} on the multi-relation benchmark, same clips as \cref{tab:preference}. Values above $0.5$ favour \ours. Human: fraction of answers preferring \ours, $20$ participants.}
  \label{tab:human}
\end{table}

\subsection{Video examples}
\label{app:video}
\begin{figure*}[t]
  \centering
  \includegraphics[width=1\linewidth]{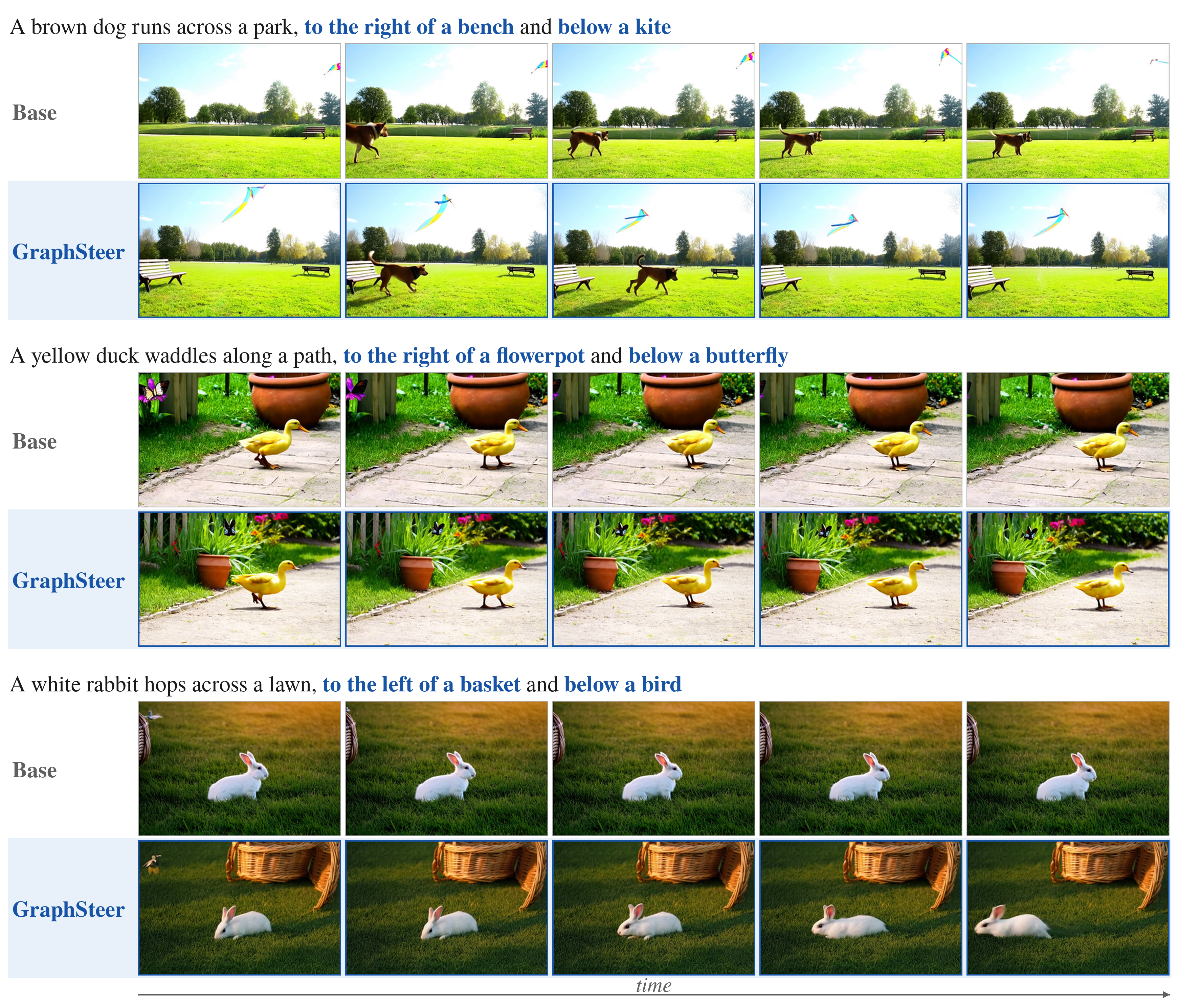}
  \caption{Base vs.\ \ours on CogVideoX-5b, with the same prompt and seed.}
  \label{fig:teaser_full}
\end{figure*}

Here we report additional video examples generated by our method and the base model (\cref{fig:teaser_full}).

\subsection{Transfer to image generation}
\label{app:images}
We also test whether our approach transfers to text-to-image models. We train one \gmn for SD2.1 and one for FLUX.1-dev, and steer prompts whose relations form a chain or a star over three or four objects ($90$ prompts $\times$ $3$ seeds, scored with GroundingDINO). We compare against Learn-to-Steer \cite{learn2steer}, Attend-and-Excite \cite{chefer2023attendexcite}, and LMD+ \cite{lian2023llm}, which plans boxes with an LLM.

\begin{table}[t]\centering\small
  \setlength{\tabcolsep}{4pt}
  \resizebox{\linewidth}{!}{%
  \begin{tabular}{lcc}
    \toprule
    method & per-edge & all-rel. \\
    \midrule
    \multicolumn{3}{l}{\emph{SD2.1}} \\
    \quad base                       & 0.12 & 0.00 \\
    \quad Attend-and-Excite \cite{chefer2023attendexcite} & 0.13 & 0.01 \\
    \quad LMD+ (planned boxes, SD1.x) \cite{lian2023llm} & 0.33 & 0.08 \\
    \quad L2S \cite{learn2steer}     & 0.41 & 0.18 \\
    \oursrow\quad \ours                      & \best{0.58} & \best{0.37} \\
    \midrule
    \multicolumn{3}{l}{\emph{FLUX.1-dev}} \\
    \quad base                       & 0.24 & 0.03 \\
    \quad L2S \cite{learn2steer}     & 0.52 & 0.24 \\
    \oursrow\quad \ours                      & \best{0.60} & \best{0.32} \\
    \bottomrule
  \end{tabular}}
  \caption{\textbf{Multi-relation image steering} ($90$ prompts $\times$ $3$ seeds per
    backbone). L2S uses its released classifiers, and LMD+ its SD1.x release with boxes from an open
        planner. \ours is the best across two backbones.}
  \label{tab:images}
\end{table}

\begin{figure}[t]
  \centering \setlength{\tabcolsep}{1pt}
  \begin{tabular}{ccc}
    \includegraphics[width=.325\linewidth]{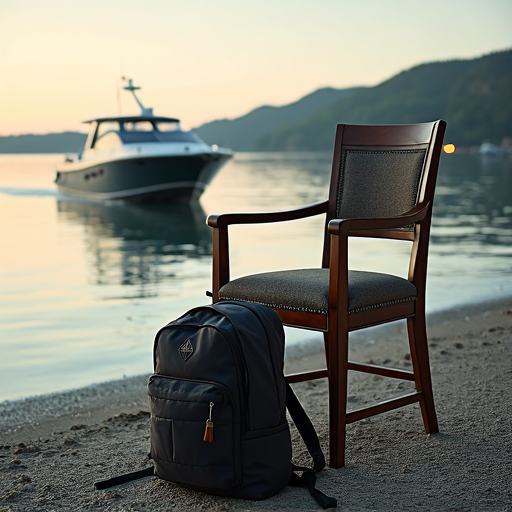} &
    \includegraphics[width=.325\linewidth]{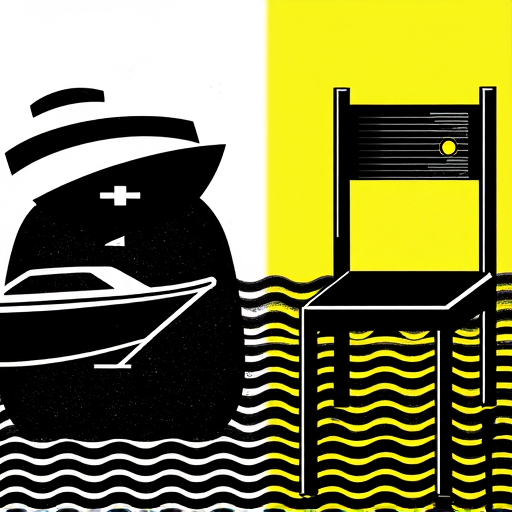}  &
    \includegraphics[width=.325\linewidth]{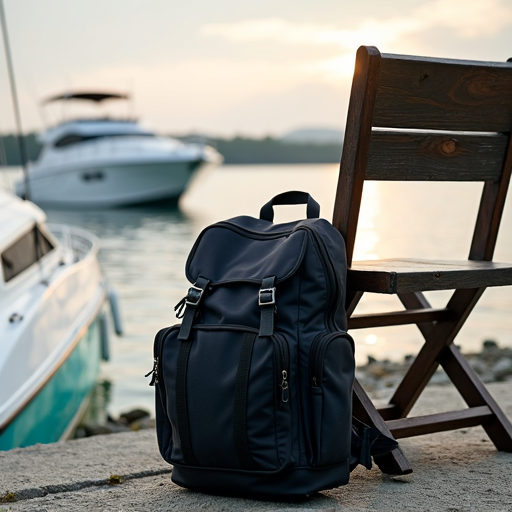}  \\
    \small base & \small L2S \cite{learn2steer} & \small \ours \\
  \end{tabular}
  \caption{FLUX.1-dev, same seed, prompt ``a boat to the left of a backpack, and a backpack to the
    left of a chair''. Base breaks the order and L2S drops the backpack; only \ours satisfies both
    relations.
    }
  \label{fig:img_qual}
\end{figure}

\begin{figure}[t]
  \centering
  \includegraphics[width=\linewidth]{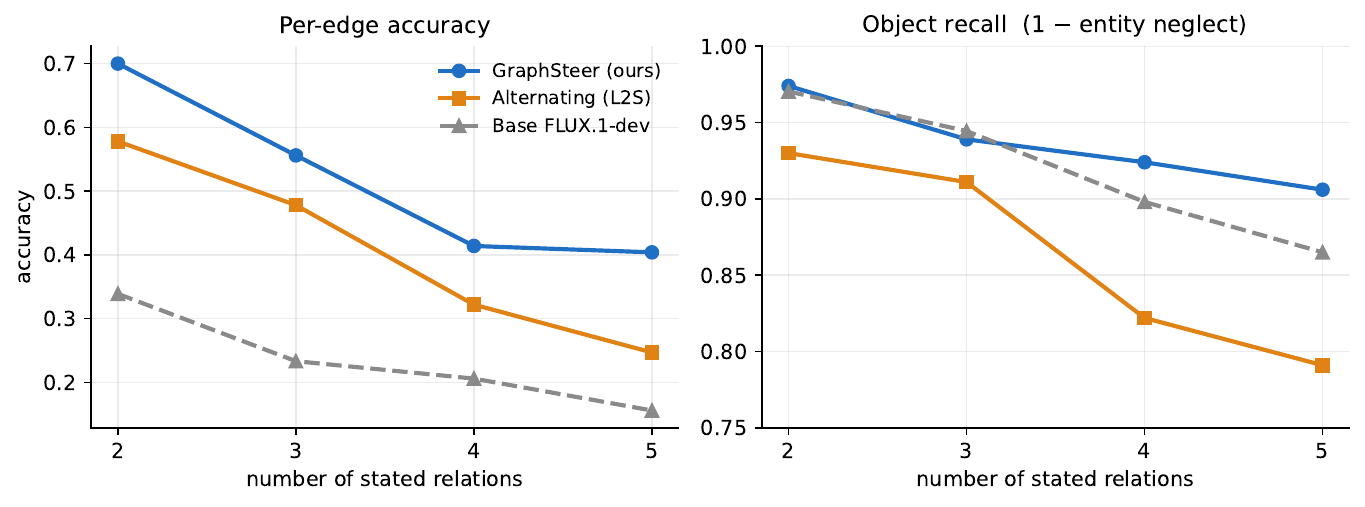}
  \caption{\textbf{More relations} on FLUX.1-dev, with chains of $3$ to $6$ objects. Left: per-edge
    accuracy. Right: object recall.}
  \label{fig:scaling}
\end{figure}

\paragraph{Results.}
\ours gives the best per-edge accuracy and all-relations rate on both backbones, ahead of LMD+, which plans boxes, and of L2S, which steers one relation at a time (\cref{tab:images}). \cref{fig:img_qual} shows a typical case: base breaks the order of the objects and L2S drops one of them. As chains grow from three to six objects (\cref{fig:scaling}), per-edge accuracy falls for every method, but the gap between \ours and L2S widens to $+0.16$, and \ours keeps almost all objects ($0.91$ recall with six objects) while L2S starts to drop them. The ternary-relation study of \cref{app:between} uses the same SD2.1 setup.

\section{Baselines}
\label{app:baselines}
\subsection{Running the competitors}
\label{app:competitors}
We run each competitor as close to its authors' setting as its release allows, and check every component we had to change. VideoTetris and TTOM run from their released code at the released settings, with only the closed-source planners replaced by open Qwen models. LVD was released for UNet backbones, so we port its released energy to the DiT backbones (``adapted''). Comp-Attn has no public code, so we reimplement it from the paper (``reimplemented''). Within each backbone, LVD and Comp-Attn share the prompts, seed and initial noise of \ours.

\paragraph{VideoTetris.} VideoTetris releases no planner, so Qwen2.5-VL-7B writes its layouts from the prompt. These layouts satisfy $98.3\%$ of the requested relations, so the planner does not limit this baseline. The T2V-CompBench scores of VideoTetris in \cref{tab:competitors} are the published ones; our runs are used only on the multi-relation benchmark and in the preference studies.

\paragraph{LVD.} We use the released code \cite{lian2024llmgroundedvideo} with all its released hyperparameters (energy loss with scale $5$, background weight $4$, threshold $200$, $5$ iterations, first $25\%$ of steps) and LVD's own layout template, planned by Qwen2.5-72B. A DiT has no UNet blocks, so the energy reads the video-to-text attention of five blocks centred on the block where our probe finds attention localizes objects best (CogVideoX $13$--$17$, Wan2.2 $5$--$9$), the most favourable choice we know of. The latent update is scaled by $\sqrt{1-\bar\alpha_t}$, or by the noise level $\sigma_t$ under flow matching. On Wan2.2 our port matches and slightly exceeds the LVD run reported by Comp-Attn ($+0.051$ vs.\ $+0.036$ spatial), and it is the strongest method there in \cref{tab:competitors,tab:multirel}. On CogVideoX, the background term at the released weight suppresses some objects (count $0.534 \to 0.432$ in \cref{tab:multirel}), which lowers its scores; we keep the released weight rather than tune it on the test prompts. Published DiT ports of LVD on CogVideoX also disagree with each other, from $+0.073$ \cite{zhang2025compattn} to $+0.010$ \cite{qu2026ttom} spatial.

\paragraph{Comp-Attn.} No code is released, so we reimplement SCI and LAM from the paper with all its stated settings (saliency temperature $0.2$, first $20\%$ of steps) and its planning prompt. The only change is the planner, Qwen2.5-72B in place of GPT-4o-mini, and every one of the $520$ prompts receives a valid layout. Our reimplementation improves both backbones and is the strongest layout method on CogVideoX in \cref{tab:competitors}. Its gains are smaller than the published ones ($+0.057$ vs.\ $+0.119$ spatial on CogVideoX, $+0.015$ vs.\ $+0.063$ on Wan2.2). Part of this comes from the base model: our CogVideoX base already scores $0.545$, against $0.517$ in the Comp-Attn paper, so in absolute terms the gap is $0.601$ vs.\ $0.636$. The rest may come from implementation details the paper does not specify.

\paragraph{TTOM.} (i) \emph{Released layouts.} With its shipped GPT-4o layouts, the released code reproduces the published T2V-CompBench spatial score ($0.632$ vs.\ $0.628$); this is the TTOM row of \cref{tab:competitors}. (ii) \emph{Planner.} TTOM's own planning code run with Qwen2.5-72B scores close to GPT-4o layouts ($0.906$ vs.\ $0.915$). We use the 72B model because the 7B model reaches only $0.71$. (iii) \emph{End to end}, the planner swap leaves the spatial score unchanged ($0.634$, $+0.002$, $95\%$ CI $[-0.04,+0.04]$), with $0.873$ layout fidelity on our benchmark. (iv) \emph{Numeracy.} TTOM ships its GPT-4o numeracy layouts in an older format that its released code cannot read. We convert them without any LLM: the boxes are unchanged, the original prompt serves as generation prompt because that format has no enriched prompt, and instance ids are renumbered so that each object's boxes are pooled as in TTOM's loss. This gives $0.478$, below the published $0.588$. Fresh layouts from TTOM's planning code with Qwen2.5-72B give about the same score ($0.493$), although that planner gets the requested number of instances right less often ($91/200$ prompts vs.\ $139/200$), so the gap does not come from the planner or the conversion. We could not close it with the released code; \cref{tab:competitors} reports our run ($0.478$), and the published score is $0.588$.

\subsection{Computational cost}
\label{app:baseline_costs}
We report computational cost in GPU-minutes, including
model loading and layout planning where applicable.
All measurements use A100-64GB GPUs.

For VideoTetris, layout planning with Qwen2.5-VL-7B
accounts for $0.07$ of the total $1.0$ GPU-min per clip.
For TTOM, planning with Qwen2.5-72B on four GPUs
accounts for $12$ of the $47$ GPU-min, while the remaining
$35$ correspond to generation with a per-video LoRA.

For LVD and Comp-Attn, we use Qwen2.5-72B on three GPUs
for layout planning, costing $4.9$ and $2.2$ GPU-min
per prompt, respectively. Both methods run on the same
backbones and seeds as \ours.

On CogVideoX-5b, Comp-Attn's LAM modifies the joint
text-video attention, preventing the use of fused attention
kernels during the first $20\%$ of denoising steps.
This introduces additional computational overhead
compared to standard sampling.

\subsection{Statistical reporting}
\label{app:statistical_reporting}
Following prior work on compositional video generation,
including LVD~\cite{lian2024llmgroundedvideo},
VideoTetris~\cite{tian2024videotetris},
TTOM~\cite{qu2026ttom}, and
Comp-Attn~\cite{zhang2025compattn},
as well as T2V-CompBench~\cite{sun2024t2vcompbench},
which evaluates one video per prompt, we report one run
with a fixed seed per prompt in
\cref{tab:competitors,tab:multirel,tab:preference,tab:transfer}.

The main constraint is computational cost. Summing the
per-clip costs in \cref{tab:multirel} across the eight
configurations we run (base, LVD, Comp-Attn and GraphSteer
on both CogVideoX-5b and Wan2.2-A14B) gives about
123~GPU-minutes per prompt. An additional seed on the
multi-relation benchmark alone (120 prompts) would
therefore cost about 245 A100 GPU-hours, with several
times more required for T2V-CompBench.

For fair comparisons, our runs of base, LVD, Comp-Attn
and GraphSteer share the same prompt, seed, and initial
noise within each backbone. Published scores for VideoTetris and ViCo in
\cref{tab:competitors} follow their original protocols.
TTOM uses its own prompt enrichment and seed.

\section{Hyperparameters}
\label{app:hyperparamters}

\cref{tab:hparams} lists every hyperparameter of training and guidance. The training maps come from CogVideoX-5b: each clip is forward-noised to step $5$ and passed once through the model, with the true prompt and with a corrupted one (\cref{sec:classifier}). Horizontal flips swap left and right in the labels, and time reversal swaps toward and away. The other augmentations simulate other backbones (\cref{app:reader}). We select the checkpoint with the best balanced accuracy on the validation split. The guidance settings are the same on every backbone and every benchmark.

\subsection{Backbones and extraction blocks}
\label{app:backbones}
\cref{tab:backbones} presents the generation settings for each video backbone and specifies the block utilized by the \gmn. All backbones are executed for 50 denoising steps with seed 13, and both base and steered clips are initialized with identical noise. The \gmn is trained exclusively on CogVideoX-5b. For the remaining backbones, the same checkpoint is loaded, with only the block name for attention reading modified. For each entity, attention is extracted from video queries to the entity's text tokens. In Wan, this corresponds to the cross-attention layer of the block. CogVideoX and the dual-stream blocks of HunyuanVideo attend to both video and text tokens; however, only the video-to-text component is retained in this context. In HunyuanVideo, attention is read from the 15th of its 20 dual-stream blocks, which precede its 40 single-stream blocks. Wan2.2-A14B contains two experts, and during the 8 guided steps, the high noise expert is used; therefore, only that expert's block is accessed.

\begin{table*}[t]\centering\small
  \setlength{\tabcolsep}{4pt}
  \resizebox{\textwidth}{!}{%
  \begin{tabular}{lcccccl}
    \toprule
    backbone & frames & resolution & latent grid ($F{\times}H{\times}W$) & CFG & blocks / heads & block read by the \gmn \\
    \midrule
    CogVideoX-5b \cite{yang2024cogvideox} & $49$ & $480{\times}720$ & $13{\times}30{\times}45$ & $6.0$ & $42$ / $48$ & \texttt{transformer\_blocks.15.attn1} (joint attention) \\
    Wan2.2-T2V-A14B \cite{wan2025} & $81$ & $480{\times}832$ & $21{\times}30{\times}52$ & $4.0$ / $3.0$ & $40$ / $40$ per expert & \texttt{blocks.15.attn2} (cross-attention, high-noise expert) \\
    Wan2.1-T2V-14B \cite{wan2025} & $81$ & $480{\times}832$ & $21{\times}30{\times}52$ & $5.0$ & $40$ / $40$ & \texttt{blocks.15.attn2} (cross-attention) \\
    HunyuanVideo \cite{kong2024hunyuanvideo} & $61$ & $480{\times}848$ & $16{\times}30{\times}53$ & $6.0$ (embedded) & $20{+}40$ / $24$ & \texttt{transformer\_blocks.15.attn} (dual-stream joint attention) \\
    \bottomrule
  \end{tabular}}
  \caption{\textbf{Video backbones.} Generation settings and the block read by the \gmn. Block names follow the Diffusers implementation of each model. Wan2.2-A14B uses one guidance scale per expert (high / low noise). HunyuanVideo has $20$ dual-stream and $40$ single-stream blocks. All runs use $50$ steps and seed $13$.}
  \label{tab:backbones}
\end{table*}

\begin{table}[t]\centering\small
  \setlength{\tabcolsep}{4pt}
  \resizebox{\linewidth}{!}{%
  \begin{tabular}{ll}
    \toprule
    hyperparameter & value \\
    \midrule
    \multicolumn{2}{l}{\emph{\gmn training}} \\
    training backbone & CogVideoX-5b, block $15$, $48$ heads \\
    capture step & $t{=}5$ of $50$, forward noising \\
    canonical map grid & $8{\times}16{\times}16$, mass-normalized \\
    encoder & $2$ layers, $d{=}256$, $4$ heads, FFN $512$, dropout $0.1$ \\
    graph network & $2$ graph-attention layers, $4$ heads, hidden $256$ \\
    edge heads & one softmax per axis \\
    optimizer & AdamW, lr $10^{-4}$, weight decay $0.01$ \\
    schedule & $128$k iterations, $6.4$k linear warm-up, cosine decay \\
    gradient accumulation / clipping & $8$ / $1.0$ \\
    label smoothing & $0.1$ \\
    precision & bf16 \\
    validation split & $15\%$, evaluated every $3.2$k iterations \\
    early stopping & patience $8$ evaluations, balanced accuracy \\
    augmentations & h-flip, time reversal, head subsets ($12$--$48$), \\
                  & grid jitter, zoom-out, attention gamma \\
    \midrule
    \multicolumn{2}{l}{\emph{Guidance (all backbones)}} \\
    guided steps & $0$--$7$ of $50$, one update per step \\
    step length $\eta$ (\cref{eq:update}) & $6$, normalized gradient \\
    presence weight $\lambda$ / threshold (\cref{eq:pres}) & $4$ / $0.5$ \\
    edge weight & $1$ for every edge \\
    entity parser & Qwen2.5-3B-Instruct \\
    relation parser & Qwen2.5-VL-7B-Instruct, text only, greedy \\
    \bottomrule
  \end{tabular}}
  \caption{\textbf{Hyperparameters} of \gmn training and of guidance. One checkpoint and one guidance configuration serve every video result.}
  \label{tab:hparams}
\end{table}

\subsection{Graph Scorer Module design}
\label{app:reader_design}
\cref{tab:reader_design} reports the four versions of the \gmn, trained on the same dataset. Each version steers CogVideoX-5b on the same 60 left/right prompts of T2V-CompBench spatial with one seed, evaluated using the official evaluator (\cref{tab:reader_design}). All four versions improve left/right accuracy over the base model by 0.21 to 0.25. The differences among the versions at n=60 are not statistically significant. The largest observed difference is +0.044 (step 5 versus all steps for the resolution-free encoder). We opted for the resolution-free encoder at step 5, as it does not reduce performance on CogVideoX and is compatible with Wan2.2-A14B and HunyuanVideo without modification.

\begin{table}[t]\centering\small
  \setlength{\tabcolsep}{4pt}
  \begin{tabular}{llcc}
    \toprule
    map grid & steps read & left/right$\uparrow$ & $\Delta$ vs base \\
    \midrule
    \multicolumn{2}{l}{base CogVideoX-5b} & 0.532 & -- \\
    native ($13{\times}30{\times}45$) & all captured & 0.738 & $+0.206$ \\
    native ($13{\times}30{\times}45$) & step $5$ & 0.774 & $+0.242$ \\
    resampled ($8{\times}16{\times}16$) & all captured & 0.740 & $+0.208$ \\
    \oursrow resampled ($8{\times}16{\times}16$) & step $5$ (ours) & \best{0.784} & \best{$+0.252$} \\
    \bottomrule
  \end{tabular}
  \caption{\textbf{Reader design.} CogVideoX-5b, T2V-CompBench spatial, $60$ left/right prompts, one seed, official evaluator, guidance without the presence term. Every version is trained on the same data.}
  \label{tab:reader_design}
\end{table}

\section{Limitations and societal impact}
\label{app:limits}
\subsection{Limitations}
\label{app:limitations}
The validation accuracy of the \gmn does not always predict its effectiveness during steering. Although depth relations can be recognized from attention maps, steering has limited effect on them (\cref{app:depth}). Our approach also inherits some limitations of the base generator. When prompts become too complex, the generator may fail to produce some objects, making it difficult to enforce their relations. On HunyuanVideo, steering can also change the scene composition or produce a cartoon style instead of preserving the original appearance.

\subsection{Societal impact}
\label{app:societal_impact}

\ours helps a text-to-video model place objects as the prompt describes, without a layout, a planner, or retraining of the generator. It adds $1.1$ GPU-minutes per clip on CogVideoX-5b ($4.2$ to $5.3$) and $1.8$ on Wan2.2-A14B ($17.2$ to $19.0$), less than LVD, Comp-Attn, and TTOM (\cref{tab:multirel}).

Better control over where objects appear lowers the effort needed to make fabricated or misleading footage, as with any improvement to video generators. \ours only changes the relative position of objects through the attention of an existing model. It adds no control over identity or appearance.

The \gmn is trained on VidOR, ImageNet-VidVRD, GQA, and synthetic clips (\cref{app:data}), and our benchmarks rely on Qwen models, GroundingDINO, and Depth-Anything for parsing and scoring. \ours therefore inherits the biases of the generators, of these datasets, and of these models. We did not measure them.